\documentclass{amsart}

\newtheorem{theorem}{Theorem}[section]
\newtheorem{proposition}{Proposition}[section]

\newtheorem{lemma}[theorem]{Lemma}

\theoremstyle{definition}
\newtheorem{definition}[theorem]{Definition}

\usepackage[margin=1.5in]{geometry} 
\usepackage{stackrel}

\newcommand{\expect}{{\rm I\!E}}
\newcommand{\real}{{\rm I\!R}}

\newcommand{\prob}{{\rm I\!P}}
\newcommand{\ball}{{\rm I\!B}}

\usepackage{amsmath,amsthm,amssymb,bm}
\usepackage{algpseudocode}
\usepackage{mathtools}
\usepackage{listings}
\usepackage{color} 
\usepackage{graphicx}
\usepackage{easybmat}
\usepackage{bm}
\usepackage{subfigure}
\usepackage{hyperref}
\usepackage{algorithm}
\usepackage{dsfont}
\usepackage{enumitem}
\usepackage{amsmath}
\usepackage{algpseudocode}
\usepackage{bbm}
\usepackage{dsfont}
\usepackage{epstopdf}
\usepackage{graphicx,subfigure}
\usepackage[labelfont=bf]{caption}
\usepackage[export]{adjustbox}
\usepackage{amsbsy}
\usepackage{romannum}
\usepackage{tikz}
\usepackage{lettrine}

\let\exp\relax

\DeclareMathOperator{\exp}{exp}

\DeclareMathOperator{\mix}{mix}

\let \expect \relax
\newcommand{\expect}{{\rm I\!E}}

\usepackage{filecontents}
\usepackage{mathrsfs}

\newcommand{\df}{\stackrel{\text{def}}{=}}

\theoremstyle{remark}

\newcommand{\qedw}{\tag*{$\square$}}
\newcommand{\qedb}{\tag*{$\blacksquare$}}

\DeclareMathOperator{\relint}{relint}

\usepackage{hyperref}
\hypersetup{
	colorlinks=true,
	linkcolor=[RGB]{110,25,25},
	citecolor=[RGB]{25,25,110},
	filecolor=magenta,      
	urlcolor=black,
}
\usepackage{lipsum}

\newcommand\blfootnote[1]{%
	\begingroup
	\renewcommand\thefootnote{}\footnote{#1}%
	\addtocounter{footnote}{-1}%
	\endgroup
}

\numberwithin{equation}{section}

\begin{document}
	
	\title[Stochastic Primal-Dual Methods for Learning Mixture Policies in MDPs]{On Sample Complexity of Projection-Free Primal-Dual Methods for Learning Mixture Policies in Markov Decision Processes}

	
	

	\author[Masoud Badiei Khuzani, et al.]{Masoud Badiei Khuzani$^{\dagger,\ddagger}$, Varun Vasudevan$^{\mathsection}$, Hongyi Ren$^{\ddagger}$, Lei Xing$^{\ddagger}$}

\address{Stanford University, 450 Serra Mall, Stanford, CA 94305\\
}
\curraddr{}
\email{mbadieik,devan,hongyi,lei@stanford.edu}

\begin{abstract}
		We study the problem of learning policy of an infinite-horizon, discounted cost,
		Markov decision process (MDP) with a large number of states. We compute the actions of a policy that is nearly as good as a policy chosen by a suitable oracle from a given mixture policy class characterized by the convex hull of a set of known base policies. To learn the coefficients of the mixture model, we recast the problem as an approximate linear programming (ALP) formulation for MDPs, where the feature vectors correspond to the occupation measures of the base policies defined on the state-action space. We then propose a projection-free stochastic primal-dual method with the Bregman divergence to solve the characterized ALP. Furthermore, we analyze the probably approximately correct (PAC) sample complexity of the proposed stochastic algorithm, namely the number of queries required to achieve near optimal objective value. We also propose a modification of our proposed algorithm with the polytope constraint sampling for the smoothed ALP, where the restriction to lower bounding approximations are relaxed. 
		In addition, we apply the proposed algorithms to a queuing problem, and compare their performance with a penalty function algorithm. The numerical results illustrates that the primal-dual achieves better efficiency and low variance across different trials compared to the penalty function method.
\end{abstract}
	\maketitle

\section{Introduction}
\blfootnote{\footnotesize{\vspace{-4mm}\\  $^{\dagger}$Department of Management Science, Stanford University, CA 94043, USA.\\
		$^{\ddagger}$Department of Radiation Oncology,	Stanford University, CA 94043, USA.\\
		$^{\mathsection}$∗Institue for Computational and Mathematical Engineering, Stanford University, CA 94043, USA.\\
	This manuscript is the extended paper accepted to 58th Conference on Decision and Control (CDC).}}
\lettrine{\textbf{M}}{arkov decision processes} are mathematical models for sequential decision making when outcomes are uncertain. The Markov
decision process model consists of decision epochs, states, actions, transition probabilities, and
costs. Choosing an action in a state generates a cost and determines the state at the next
decision epoch through a transition probability function. Policies or strategies are prescriptions
of which action to take in a given state to minimize the cost. Given a MDP, the main objective is to compute (near-) optimal policies that (approximately) attain the minimum long-term average cost.

In most practical applications, the underlying MDP is compactly represented and the number of states scale exponentially with the size of the representation of the MDP. In addition, for such applications, various hardness results often indicate that computing actions of optimal policies is intractable in the sense that polynomial-time algorithms to compute such control policies are unlikely (unless come complexity class collapse) or simply don't have guarantees for searching policies within a constant multiplicative or additive factor of the optimal; see, \textit{e.g.},  \cite{banijamali2018optimizing}. In view of those negative results, it is natural to pursue a more modest objective which is to compute the actions of a policy that is nearly as good as a policy chosen by an oracle from a given restricted policy class. Following the work of \cite{banijamali2018optimizing}, in this paper we consider the policy class to be the convex-hull of a set of known base policies. Such base policies are often known in practice for certain applications. For instance, a myopic and a look ahead policy in queuing networks can be combined to achieve a mixture policy with better performance.

\subsection{Main Contributions}

The main contributions of this paper are summarized as follows:
\begin{itemize}
	\item We formulate the optimization over the restricted class of mixture policies as an approximate linear programming (ALP) for MDPs, where the feature vectors are given by the occupation measures of the base policies. 
	
	\item We propose a novel \textit{projection-free} stochastic primal-dual (SPD) algorithm for the reinforcement learning of efficient mixture policies in Markov decision processes.
	
	\item We analyze the constraint violation of the solutions of SPD, and prove that such constraint violation diminishes in the asymptotic of many rounds.
	
	\item We analyze the sample complexity of the proposed algorithm, \textit{i.e.}, the number of queries required from a sampling oracle to achieve near optimal cost function.
	
	\item We numerically compare the performance of the proposed primal-dual algorithm with that of the penalty function method for a queuing problem, and we show that the solutions obtained by the proposed method in this paper has a smaller variance across different trials compared to the penalty function method.
\end{itemize}

\subsection{Connections to Prior Works}
The ALP as a framework to find a ``low-dimensional" representation of ``high-dimensional" functions on a state (action) space has a long history in decision theory; see, \textit{e.g.},  \cite{schweitzer1985generalized,de2003linear,chen2018scalable,abbasi2014linear,roy2003approximate} and references therein. The seminal work of De Farias and Von Roy \cite{de2003linear} studies an ALP for stochastic control problems as a means of alleviating the curse of dimensionality. The main challenge in using the proposed ALP is that while it has a relatively small number of variables, it has an intractable number of constraints. To address this issue, the same authors proposed a constraint sampling scheme in a separate work \cite{de2004constraint}.  In the same line of work, a dual approximate linear programming for the MDP is considered in \cite{abbasi2014linear}, where the optimization variable is a stationary distribution over state-action pairs. A neighborhood of a low-dimensional subset of the set of stationary distributions defined in terms of state-action features is considered as the comparison class. In a similar line of work, a $\pi$-learning algorithm is proposed in \cite{chen2018scalable} which leverages state and action features to reduce the dimensionality of the state and action spaces. In addition, the sample complexity of the proposed $\pi$-learning algorithm is analyzed. 

Our work is also closely related to the recent study of Banijamali, \textit{et al.} \cite{banijamali2018optimizing}. Therein, the authors propose a stochastic gradient decent in conjunction with a penalty function method to optimize over a set of mixtures of policies. The main challenge in using a penalty function method is that its performance is often sensitive to the choice of the penalty factor in addition to the learning rate. Moreover, it yields a large constraint violation as is observed in \cite[Thm. 1]{mahdavi2012trading}. Furthermore, to optimize the regret performance , the authors in \cite{banijamali2018optimizing} propose a penalty factor that depends on the amount of violation of constraints, which is unknown in practice. In this paper, we propose a stochastic primal-dual method whose only hyper-parameter is the learning rate. 

We also mention the recent work of Chen and Wang \cite{chen2018scalable}, where a primal-dual algorithm with the Bregman divergence is considered in conjunction with the approximate linear programming formulation of MDPs. The work of \cite{chen2018scalable} deals with the \textit{sample complexity} of the stochastic primal-dual algorithm, namely the number of queries to the oracle needed to attain near optimal value function, whereas in this paper we are concerned with the \textit{efficiency} in the dual space as well as the \textit{constraint violation} of the primal-dual  solutions. In addition, the algorithm we propose differs from \cite{chen2018scalable} in several key points. First, unlike the algorithm of Chen and Wang, our approach does not require the \textit{realizability} assumption (cf. \cite[Def. 1]{chen2018scalable}). The realizability condition requires the spanning vectors of features in ALP to be non-negative, which in turn allows one to simplifies the projection onto the simplex section of a hyper-plane. In particular, projection onto the simplex can be implemented efficiently using the Kullback-Leibler (KL) divergence as the proximal function. In our algorithm, we provide a direct method of projection onto the hyper-plane which obviates the realizability condition and provides a more expressive representation. In addition, we present a randomized policy based on the primal-dual solutions that differs from the policy proposed in \cite{chen2018scalable}. Second, our proposed algorithm solves an optimization problem in the dual space, where feature vectors are the occupation measures of a set of base policies. This allows us to compute a randomized policy directly from the solution of the underlying dual optimization problem. Lastly, the role of the Bregman divergence in our algorithm is to implicitly enforce the constraints due to the size of the policy class. In contrast, the Bregman divergence in \cite{chen2018scalable} is used as a mean to attain better sample complexities via adaptation to the geometry of the feasible set.

\subsection{Paper Outline}

The rest of this paper is organized as follows. In Section \ref{Section:Preliminaries of Markov Decision Processes}, we present preliminaries related to MDPs. In addition, we review the approximate linear programming formulation of MDPs based on the linear representation of large state-spaces. In Section \ref{Section:Main_Results},  we present a stochastic regularized primal-dual algorithm to compute the coefficients of the mixture policy. We also present the main theorem concerning the performance of the proposed algorithm. In Section \ref{Section:Proof_of_Main_Results} we present the proof of the main theorem, while deferring technical lemmas to the appendices. Lastly, in Section \ref{Section:Conclusion and Future Works}, we conclude this paper.

\section{Preliminaries}
\label{Section:Preliminaries of Markov Decision Processes}

In this section, we first present the notations and technical definitions that we need to establish our theoretical results. We then review relevant definitions regarding infinite horizon discounted Markov decision processes. We also review Bellman's equations as well as linear programming dual formulation of Bellman's equations. 

\textit{Notations and Definitions}. We denote vectors by the lower case letters (\textit{e.g.} $\bm{x}$), and random variables and matrices with the upper case letters (\textit{e.g.} $\bm{X}$). The dual norm $\|\cdot\|_{\ast}$ conjugate to the norm $\|\cdot\|$
is defined by $\|\bm{z}\|_{\ast}\df \sup_{\|\bm{x}\|\leq 1}\langle \bm{z},\bm{x} \rangle$. We denote the Euclidean projection by $\mathcal{P}_{\mathcal{X}}(\bm{x})\df \min_{\bm{y}\in \mathcal{X}} \|\bm{x}-\bm{y}\|_{2}^{2}$. We use the standard asymptotic notation $\Omega$ with the following definition: If $f,g$ are two functions from $\mathbb{N}$ to $\mathbb{N}$, then $f = \mathcal{O}(g)$ if there exists a constant $c$ such that $f(n)\leq c · g(n)$ for every sufficiently large $n$ and that $f=\Omega(g)$ if $g =\mathcal{O}(f)$. For positive integer $m$, we use the shorthand $[m]$ to denote the set $\{1, 2, . . . , m\}$. For a matrix $\bm{X}$, let $\|\bm{X}\|_{\ast,p}=\sup_{\bm{x}\not = 0}\|\bm{\Phi}\bm{x}\|_{\ast}/\|\bm{x}\|_{p}$ denote the subordinate norm for $p\in [1,\infty)$. We denote the largest singular value by $\sigma_{\max}(\bm{X})$. Further, we use the shorthand notations $x\vee y\df \max\{x,y\}$, $x\wedge y\df \min\{x,y\}$, and $[x]_{+}\df x\vee 0$. A function $f$ is $L$-Lipschitz with respect to the norm $\|\cdot\|$ over $\mathcal{X}$ iff
\begin{align*}
|f(\bm{x})-f(\bm{y})|\leq L\|\bm{x}-\bm{y}\|, \quad \text{for all}\ \bm{x},\bm{y}\in \mathcal{X}.
\end{align*}
A function $f$ is $\beta$-smooth with respect to the norm $\|\cdot\|$ over $\mathcal{X}$ iff
\begin{align*}
\|\nabla f(\bm{x})-\nabla f(\bm{y})\|_{\ast}\leq \beta \|\bm{x}-\bm{y}\|, \quad \text{for all} \ \bm{x},\bm{y}\in \mathcal{X}. 
\end{align*}
A function $f$ is $\mu$-strongly convex with respect to the norm $\|\cdot\|$ over $\mathcal{X}$ iff
\begin{align*}
f(\bm{x})+\langle \bm{g},\bm{y}-\bm{x}\rangle +\dfrac{\mu}{2}\|\bm{x}-\bm{y}\|^{2}\geq f(\bm{y}),
\end{align*} 
for all $\ \bm{x},\bm{y}\in \mathcal{X}, \bm{g}\in \partial f(\bm{x})$. The effective domain of a function $f:\mathcal{X}\rightarrow \real$ is the following set $\text{dom}(f) \df \{\bm{x}\in \mathcal{X}:f(\bm{x})<+\infty \}$. The sub-differential set of a function at the point $\bm{x}_{0}\in \mathrm{dom}(f)$ is defined as follows
\begin{align}
\partial f(\bm{x}_{0})\df \{\bm{g}:f(\bm{x})-f(\bm{x}_{0})\geq \langle \bm{g},\bm{x}-\bm{x}_{0} \rangle, \forall \bm{x}\in \mathrm{dom}(f)\}.
\end{align}

 The relative interior of a convex set $C$, abbreviated $\relint(C)$, is defined as $\relint(C)\df \{\bm{x}\in C:\exists \epsilon, \ball_{\epsilon}(\bm{x})\cap \text{aff}(C)\subseteq C \}$, where $\text{aff}(C)$ denotes the affine hull of the set $C$, and $\ball_{\epsilon}(\bm{x})$ is a ball of radius 
$\epsilon$ centered on $\bm{x}$.

The \textit{Fenchel conjugate} of the function $f:\mathcal{X}\rightarrow \real$ is defined as follows
\begin{align}
f^{\ast}(\bm{y})=\sup_{\bm{x}\in \mathcal{X}}\{\langle \bm{x},\bm{y} \rangle-f(\bm{x})\}.
\end{align}
\begin{definition}\textsc{(Orlicz Norm)}
	\label{Def:Orlicz}
	The Young-Orlicz modulus is a convex non-decreasing function $\psi:\real_{+}\rightarrow \real_{+}$ such that $\psi(0)=0$ and $\psi(x)\rightarrow \infty$ when $x\rightarrow \infty$. Accordingly, the Orlicz norm of an integrable random variable $X$ with respect to the modulus $\psi$ is defined as
	\begin{align}
	\|X\|_{\psi}\df \inf \{\beta>0:\expect[\psi(|X|-\expect[|X|]/\beta)]\leq 1\}.
	\end{align}
\end{definition}

In the sequel, we consider the Orlicz modulus $\psi_{\nu}(x)\df \exp(x^{\nu})-1$ . Accordingly, the cases of $\|\cdot\|_{\psi_{2}}$ and $\|\cdot\|_{\psi_{1}}$ norms are called the sub-Gaussian and the sub-exponential norms and have the following alternative definitions:
\begin{definition}\textsc{(Sub-Gaussian Norm)}
	The sub-Gaussian norm of a random variable $Z$, denoted by $\|Z\|_{\psi_{2}}$, is defined as
	\begin{align}
	\|Z\|_{\psi_{2}}= \sup_{q\geq 1} q^{-1/2}(\expect|Z|^{q})^{1/q}.
	\end{align}
	For a random vector $\bm{Z}\in \real^{n}$, its sub-Gaussian norm is defined as 
	\begin{align}
	\|\bm{Z}\|_{\psi_{2}}=\sup_{\bm{x}\in \mathrm{S}^{n-1}}\|\langle  \bm{x},\bm{Z}\rangle \|_{\psi_{2}}.
	\end{align}
\end{definition}

\begin{definition}\ \textsc{(Sub-exponential Norm)}
	The sub-exponential norm of a random variable $Z$, denoted by $\|Z\|_{\psi_{1}}$, is defined as follows
	\begin{align}
	\|Z\|_{\psi_{1}}=\sup_{q\geq 1} q^{-1}(\expect[|Z|^{q}])^{1/q}.
	\end{align}
	For a random vector $\bm{Z}\in \real^{n}$, its sub-exponential norm is defined as 
	\begin{align}
	\|\bm{Z}\|_{\psi_{1}}= \sup_{\bm{x}\in \mathrm{S}^{n-1}} \|\langle \bm{Z},\bm{x}\rangle\|_{\psi_{1}}.
	\end{align}
\end{definition}

\begin{definition}\ \textsc{(Legendre Function)} A function $\phi:C\rightarrow \real$ , $C=\text{dom}(\phi)$ is called a Legendre function (\textit{a.k.a.} essentially smooth functions) if it satisfies the following conditions:
	\begin{itemize}
		\item $C\subseteq \real^{d}$ and $C\not=\emptyset$ and $\relint(C)$ is convex.
		\item $\phi$ is strictly convex.
		\item partial derivatives of $\partial\phi\over \partial x_{i}$ exists and are continuous for all $i = 1, .., d$.
		\item Any sequence $(x_{n})_{n\in \mathbb{N}}$ converging to a boundary point of $C$ satisfies
		\begin{align}
		\nonumber
		\lim_{n\rightarrow \infty}\|\nabla\phi(x_{n})\|=\infty.
		\end{align}
	\end{itemize}
\end{definition}

\begin{definition}\textsc{(Bregman Divergence)}
	\label{Def:Bregman_divergence}
	Suppose $\phi:C\rightarrow \real$ , $C=\text{dom}(\phi)$ is a Legendre function. The Bregman divergence $D_{\phi}:C\times \relint(C)\mapsto [0,\infty)$ is defined as follows
	\begin{align*}
	D_{\phi}(\bm{x};\bm{y})\df \phi(\bm{x})-\phi(\bm{y})-\langle \bm{x}-\bm{y},\nabla \phi(\bm{y})\rangle,
	\end{align*}
	where $\nabla \phi(\bm{y})$ is the gradient vector $\phi$ evaluated at $\bm{y}$.
\end{definition}

\subsection{Markov Decision Processes (MDPs)}
\label{Eq:Prelim_SubSectionA}
In this paper, we consider MDPs with high dimensional state and action spaces. The first definition formalizes MDPs:
\begin{definition}
	\textsc{(Markov Decision Process)} A Markov decision process (MDP) is a 6-tuple $(T,\mathcal{S},\mathcal{A},P,c,\gamma)$ consists of:
	\begin{itemize}
		\itemsep0em 
		\item \textsc{Decision epochs}: The set $\{0,1,2,\cdots,T-1\}$ represents the set of times at
		which decisions are to be made. If $T$ is finite, then the MDP is said to be a finite
		horizon MDP with $T$-epochs. If $T=\infty$, the MDP is said to be an infinite
		horizon MDP. 
		\item \textsc{States}: We assume that the state-space $\mathcal{S}=\{1,2,\cdots,n\}$ is finite.
		\item \textsc{Actions}: We assume that the action set $\mathcal{A}=\{1,2,\cdots,m\}$ is also finite.
		\item \textsc{Transition model}: The transition model $P_{a}(\cdot,s)$ for each $s\in \mathcal{S}$ and $a\in \mathcal{A}$, is the probability distribution $P_{a}(\cdot,s)$ on $\mathcal{S}$. The element $P_{a}(s',s)$ represents the probability
		of transitioning to the state $s'\in \mathcal{S}$ after performing the action $a\in \mathcal{A}$ in the state $s\in \mathcal{S}$. We define the matrices $\bm{P}_{a}\df [P_{a}(s,s')]_{(s,s')\in \mathcal{S}\times \mathcal{S}}$ and $\bm{P}\df [P_{a}(s',s)]_{(s,a)\in \mathcal{S}\times \mathcal{A},s'\in \mathcal{S}}\in \real^{nm\times n}$.
		\item \textsc{Cost function}: For each $a\in \mathcal{A}$,  $c_{a}:\mathcal{S} \mapsto [c_{\min},c_{\max}]$  is the cost function. Let $\bm{c}_{a}\df (c_{a}(1),\cdots,c_{a}(n))$, and $\bm{c}\df (\bm{c}_{1},\cdots,\bm{c}_{m})$.
		\item \textsc{Discount factor}: The discount factor $\gamma\in(0,1)$ reflects the intertemporal preferences.
	\end{itemize}
\end{definition}
We consider discounted infinite horizon MDP characterized by the tuple $(T=\infty,\mathcal{S},\mathcal{A},\bm{P},\bm{c},\gamma)$. For such MDPs, we compute  randomized policies which we formally define below: 
\begin{definition}
\textsc{(Randomized Policy)}
	A randomized policy $\bm{\pi}$ is the sequence of distributions, where $\bm{\pi}:\mathcal{S}\rightarrow \mathcal{A},s\mapsto \bm{\pi}(s)=(\pi_{1}(s),\cdots,\pi_{m}(s))$ is a probability distribution on the action space $\mathcal{A}$, conditioned on the state $s\in \mathcal{S}$. The value $\pi_{a}(s)$ represents the probability of taking the action $a\in \mathcal{A}$ in the state $s\in \mathcal{S}$. Let $\Pi$ denotes the set of all such randomized policies.
\end{definition}

The objective of discounted infinite horizon MDP is to find a policy $\bm{\pi}\in \Pi$ such that the infinite-horizon sum of discounted costs is minimized regardless of the initial state $s_{0}\in \mathcal{S}$, \textit{i.e.},
\begin{align*}
\min_{\bm{\pi}\in \Pi}v^{\bm{\pi}}(s)=\lim\inf_{T\rightarrow \infty}\expect^{\bm{\pi}}\left[\sum_{t=0}^{T}\gamma^{t} c_{a_{t}}(s_{t})\Bigg|s_{0}=s\right],
\end{align*} 
where $s_{1},s_{2},\cdots$ and $a_{0},a_{1},\cdots$ are the realizations of state transitions and actions, respectively, generated by the Markov decision process under a given policy $\bm{\pi}$, and the expectation $\expect^{\bm{\pi}}[\cdot]$ is taken over the entire trajectory.

Define the Bellman operator $(\mathscr{S}\bm{v})(s)\df\min_{a\in \mathcal{A}}\left\{\gamma \sum_{s'\in \mathcal{S}}P_{a}(s,s')v^{\ast}(s')+c_{a}(s)\right\}$ for all $s\in \mathcal{S}$. From the theory of dynamic programming, a vector is the optimal value function to the MDP if and only if it satisfies the following Bellman fixed point equation \cite{bertsekas1995dynamic}
\small \begin{align}
\label{Eq:Bellman}
\mathscr{S}\bm{v}^{\ast}=\bm{v}^{\ast},
\end{align}\normalsize
where $\bm{v}^{\ast}=(v^{\ast}(1),\cdots,v^{\ast}(n))\in \real^{n}$ is the \textit{difference-of-value vector}. A stationary policy $\bm{\pi}^{\ast}$ is an optimal policy of the MDP if it attains the
element-wise maximization in the Bellman equation \eqref{Eq:Bellman}, \textit{i.e.}, $\bm{v}^{\bm{\pi}^{\ast}}=\bm{v}^{\ast}$.

Alternatively, the Bellman equation \eqref{Eq:Bellman} can be recast as the following linear programming problem (cf. \cite{manne1960linear}),
\begin{subequations}
	\label{Eq:primal_problem}
	\begin{align}
	\label{Eq:constrained}
	&\max_{\bm{v}\in \real_{+}^{n}}\  \bm{\alpha}^{T}\bm{v}\\
	&\text{s.t.}:(\bm{I}_{n}-\gamma\bm{P}_{a})\bm{v}-\bm{c}_{a}\succeq  0,\quad \forall a\in \mathcal{A},
	\end{align}
\end{subequations}
where $\bm{c}_{a}\df  (c_{a}(1),\cdots,c_{a}(n))$, and $\bm{\alpha}\in \Delta_{n}$ is the initial distribution over the states, and $\Delta_{n}= \{\bm{x}\in \real^{n}:0\preceq \bm{x}, \sum_{i=1}^{n}x_{i}=1\}$ is the simplex of the probability measures on the set of states $\mathcal{S}$.

To characterize the dual problem associated with the primal problem \eqref{Eq:primal_problem}, Let $\bm{\mu}_{\bm{\alpha}}^{\bm{\pi}}\df (\mu_{\bm{\alpha}}^{\bm{\pi}}(1),\cdots,\mu_{\bm{\alpha}}^{\bm{\pi}}(n))$ denotes the stationary distribution under the policy $\bm{\pi}$ and initial distribution of states $\bm{\alpha}\in \Delta_{n}$. In particular, let $\bm{P}^{\bm{\pi}}\df [P^{\bm{\pi}}(s,s')]_{(s,s')\in \mathcal{S}\times \mathcal{S}}\in \real^{n\times n}$  denotes the transition probability matrix induced by a fixed policy $\bm{\pi}$ whose matrix elements are defined as $\bm{P}^{\bm{\pi}}(s,s')\df \sum_{a\in \mathcal{A}}\pi_{a}(s)P_{a}(s,s')$. Alternatively, let $\bm{M}^{\bm{\pi}}$ be a $n\times nm$ matrix that encodes the policy $\bm{\pi}$, \textit{i.e.}, let $\bm{M}^{\bm{\pi}}_{(i,(i-1)m+1)-(i,im)}=(\pi_{a}(s_{i}))_{a\in \mathcal{A}}$, where other entries of the matrix are zero. Then, $\bm{P}^{\bm{\pi}}\df \bm{M}^{\bm{\pi}}\bm {P}$.

Furthermore, let $\bm{\mu}_{\bm{\alpha}}^{\bm{\pi}}\df (\mu_{\bm{\alpha}}^{\bm{\pi}}(1),\cdots,\mu_{\bm{\alpha}}^{\bm{\pi}}(n))$ denotes the stationary distribution of the Markov chain induced by the policy $\bm{\pi}$, \textit{i.e.}, $\bm{\mu}_{\bm{\alpha}}^{\bm{\pi}}\bm{P}^{\bm{\pi}}=\bm{\mu}_{\bm{\alpha}}^{\bm{\pi}}$, where
\begin{align}
\bm{\mu}^{\bm{\pi}}_{\bm{\alpha}}&=(1-\gamma)\bm{\alpha}\sum_{t=0}^{\infty}\gamma^{t}(\bm{P}^{\bm{\pi}})^{t}\\ 
\label{Eq:Stationary_Distribution}
&=(1-\gamma)\bm{\alpha}(\bm{I}_{n}-\gamma\bm{P}^{\bm{\pi}})^{-1}.
\end{align}
The measure $\bm{\mu}_{\bm{\alpha}}^{\bm{\pi}}$ captures the expected frequency of visits to each state when the underlying policy is $\bm{\pi}$, conditioned on the initial state being distributed according to $\bm{\alpha}\in  \Delta_{n}$. Future visits are discounted by the factor $\gamma$.

We define the \textit{occupation measure} as the vector $\bm{\xi}^{\bm{\pi}}=(\bm{\xi}^{\bm{\pi}}_{a})_{a\in \mathcal{A}}=(\xi^{\pi}_{a}(s))_{s\in \mathcal{S},a\in \mathcal{A}}\in \real^{nm}$ defined by $\bm{\xi}_{a}^{\bm{\pi}}\df \bm{\pi}_{a}\odot \bm{\mu}^{\bm{\pi}}_{\bm{\alpha}}$, where $\odot$ is the Hadamard (element-wise) vector  multiplication.. Then,
\begin{align}
\nonumber
\bm{\pi}^{\ast}&=\arg\min_{\bm{\pi}\in \Pi}\sum_{s\in \mathcal{S}} \mu_{\bm{\alpha}}^{\bm{\pi}}(s) \sum_{a\in \mathcal{A}}\pi_{a}(s)c_{a}(s)\\ \nonumber
&=\arg\min_{\bm{\pi}\in \Pi }\sum_{s\in \mathcal{S}}\sum_{a\in \mathcal{A}}\xi_{a}^{\bm{\pi}}(s)c_{a}(s)\\
&=\arg\min_{\bm{\pi}\in \Pi} \bm{c}^{T}\bm{\xi}^{\bm{\pi}}.
\end{align}
Thus, the dual problem associated with the primal problem in Eq. \eqref{Eq:constrained} has the following form
\begin{subequations}
	\label{Eq:Linear_Programming}
	\begin{align}
	&\min_{\bm{\xi}\in \real^{m n}} \bm{c}^{T}\bm{\xi}\\  \label{Eq:Constraint_1}
	&\text{s.t.}: \bm{\xi}^{T}(\bm{P}-\bm{Q})=\bm{0},\\ \label{Eq:Constraint_2}
	&\bm{0}\preceq \bm{\xi}, \quad \bm{\xi}^{T}\bm{1}=1.
	\end{align}
\end{subequations}
where $\bm{Q}\in \{0,1\}^{nm\times n}$ is a binary matrix such that the $i$-th column has $m$ ones in rows $1+(i-1)m$ to $im$, and $\bm{P}\df [P_{a}(s',s)]_{(s,a)\in \mathcal{S}\times \mathcal{A},s'\in \mathcal{S}}\in \real_{+}^{nm\times n}$. In the optimization problem in Eq. \eqref{Eq:Linear_Programming}, the constraint \eqref{Eq:Constraint_2} ensures that $\bm{\xi}\in \real^{nm}$ is a distribution, and the constraint \eqref{Eq:Constraint_1} guarantees that $\bm{\xi}$ is stationary.

Let $\bm{\xi}^{\ast}\df (\xi_{a}^{\ast}(s))_{a\in \mathcal{A}}\in \real_{+}^{nm}$ denotes the optimal solution of the linear programming problem in Eq. \eqref{Eq:Linear_Programming}. An optimal randomized optimal policy $\bm{\pi}^{\ast}$ can be characterized as follows
\begin{align}
\pi_{a}^{\ast}(s)= \dfrac{\xi^{\ast}_{a}(s)}{\mu_{\bm{\alpha}}^{\bm{\pi}^{\ast}}(s)}= \dfrac{\xi^{\ast}_{a}(s)}{\sum_{a\in \mathcal{A}}\xi^{\ast}_{a}(s)},\quad \forall a\in \mathcal{A},\forall s\in \mathcal{S}.
\end{align}
Furthermore, the optimal objective value of the linear programming problem in Eq. \eqref{Eq:Linear_Programming} corresponds to the optimal discounted-cost value under the optimal policy $\bm{\pi}^{\ast}$ and the initial distribution $\bm{\alpha}\in \Delta_{n}(\mathcal{S})$ of the states.

\subsection{Approximate Linear Programming (ALP) for the Linear Representation of Large State Spaces}
\label{Subsection:ALP}

It is well-known that the sample complexity as well as computational complexity of solving the linear programming dual problem in Eq. \eqref{Eq:Linear_Programming} scales (at least) linearly with the size of the state space $n$, rendering exact representation intractable in the face of problems of practical scale; see, \textit{e.g.}, \cite{powell2007approximate}.  Consequently, to deal with MDPs with a large state space, it is practical to map the state space to a low dimensional space, using the linear span of a small number of \textit{features}. 

Specifically, let $\bm{\Psi}=[\bm{\psi}^{1},\cdots,\bm{\psi}^{d}]\in \real^{nm\times d}$ denotes a matrix whose column vectors represent feature vectors. In the feature space, the distribution $\bm{\xi}$ is spanned by the linear combination of features $\bm{\theta}\in \bm{\Theta}_{R}^{d}\df \{\bm{\theta}\in \real^{d}:\|\bm{\theta}\|\leq R\}$, namely, $\bm{\xi}^{{\bm{\theta}}}=\bm{\Psi}\bm{\theta}$. Here, the radius $R>0$ and the general norm $\|\cdot\|$ determine the size and geometry of the policy class $\bm{\Theta}_{R}^{d}$, respectively. The dual optimization problem in Eq. \eqref{Eq:Linear_Programming} can thus be reformulated in terms of features 
\begin{subequations}
	\label{Eq:dual_formulation}
	\begin{align}
	&\min_{\bm{\theta}\in \bm{\Theta}_{R}}    \bm{c}^{T}\bm{\Psi}\bm{\theta}\\
	&\text{s.t.}: \bm{\theta}^{T}\bm{\Psi}^{T}(\bm{P}-\bm{Q})=\bm{0},\\
	&\bm{0}\preceq \bm{\Psi}\bm{\theta}, \quad \bm{\theta}^{T}\bm{\Psi}^{T}\bm{1}=1.
	\end{align}
\end{subequations}

Designing the feature matrix $\bm{\Psi}$ for Markov decision processes is a challenging problem. In particular, we wish to design a feature matrix $\bm{\Psi}$ such that the linear expansion $\bm{\Psi}\bm{\theta}$ is \textit{expressive}, but does not lead to \textit{over-fitting}. In this paper, we focus on the set of features associated with a set of known base policies. Formally, we consider the set of $d$-base policies $\bm{\pi}^{1},\cdots,\bm{\pi}^{d}\in \Pi$ and define the subset $\Pi^{d}\subset \Pi$ as the convex hull of base policies,
\begin{align}
\label{Eq:Mixture_Policies}
\Pi^{d}\df \left\{\bm{\pi}^{\bm{\omega}}:\bm{\pi}^{\bm{\omega}}=\sum_{i=1}^{d}\omega_{i}\bm{\pi}^{i},\sum_{i=1}^{d}\omega_{i}=1,\omega_{i}\geq 0, i=1,2,\cdots,d\right \}.
\end{align}

Corresponding to each base policy $\bm{\pi}^{i},i=1,2,\cdots,d$, a stationary distribution $\bm{\mu}^{i}\df \bm{\mu}^{\bm{\pi}^{i}}$ is defined according to Eq. \eqref{Eq:Stationary_Distribution}. The dual space of $\Pi^{d}$ in Eq. \eqref{Eq:Mixture_Policies} is then defined as the linear combinations of occupation measures
\begin{align}
\Xi^{d}\df \left\{\bm{\xi}^{\bm{\theta}}:\bm{\pi}^{\bm{\omega}}=\sum_{i=1}^{d}\omega_{i}\bm{\pi}^{i},\sum_{i=1}^{d}\omega_{i}=1, i=1,2,\cdots,d\right \}.
\end{align}
  
For all $i\in [d]$ the state-action distribution is defined as $\bm{\psi}^{i}_{a}\df \bm{\pi}^{i}_{a}\odot\bm{\mu}^{i}$ for all $a\in  \mathcal{A}$.  With this choice of the feature vectors, we have $\bm{\Psi}^{T}\bm{1}=\bm{1}$ and $\bm{\Psi}^{T}(\bm{P}-\bm{Q})=\bm{0}$ as the columns of the matrix $\bm{\Psi}$ are stationary probability distributions. Therefore, the dual optimization \eqref{Eq:dual_formulation} takes the following form
\begin{subequations}
	\label{Eq:Dual_MixturePolicy}
	\begin{align}
	&\min_{\bm{\theta}\in \bm{\Theta}_{R}} \bm{c}^{T}\bm{\Psi}\bm{\theta}\\
	&\text{s.t.}: \bm{0}\preceq \bm{\Psi}\bm{\theta}, \quad \bm{\theta}^{T}\bm{1}=1.
	\end{align}
\end{subequations}

Let $\bm{\Theta}^{d}_{1}\df \{\bm{\theta}\in \real^{d}:\bm{0}\preceq \bm{\Psi}\bm{\theta}\}$ and $\bm{\Theta}_{2}^{d}\df \{\bm{\theta}\in \real^{d}: \bm{\theta}^{T}\cdot\bm{1}=1 \}$. The feasible set of the dual optimization in Eq. \eqref{Eq:Dual_MixturePolicy} is the intersection of the following sets
\begin{align}
\bm{\Theta}^{d}=\bm{\Theta}_{1}^{d}\cap \bm{\Theta}_{2}^{d}\cap \bm{\Theta}_{R}^{d}.
\end{align}
Let $\widehat{\bm{\theta}}_{T}$ denotes an approximate solution of the optimization problem in Eq. \eqref{Eq:dual_formulation} generated by an (arbitrary) optimization algorithm after $T$ rounds. When $\widehat{\bm{\theta}}_{T}\in \bm{\Theta}^{d}$ is a feasible solution of Eq. \eqref{Eq:dual_formulation}, then $\bm{\xi}^{{\widehat{\bm{\theta}}_{T}}}=\bm{\Psi}\widehat{\bm{\theta}}_{T}$ defines a valid occupation measure. However, in this paper we permit the feature vectors to violate the non-negativity constraint defined by $\bm{\Theta}_{1}^{d}$ in the following sense:  let $V:\real^{d}\rightarrow \real_{+},\bm{\theta}\mapsto V(\bm{\theta})$ denotes a function that quantifies the amount by which the vector $\bm{\Psi}\widehat{\bm{\theta}}_{T}$ violates the non-negativity constraints. In particular, $V(\bm{\theta})=0$ iff $\bm{\theta}\in \bm{\Theta}_{1}^{d}$. After $T$ rounds of the algorithm we propose in the next section (cf. Algorithm \ref{Alg:1}), it outputs an estimate $\bm{\widehat{\bm{\theta}}}_{T}$ that may potentially violate the constraints defined by $\bm{\Theta}_{1}^{d}$, and thus $V(\bm{\widehat{\bm{\theta}}}_{T})\geq 0$. Nevertheless, we impose the constraint that the estimate generated by the algorithm satisfies the constraints in the asymptotic of many rounds $\lim_{T\rightarrow \infty}V(\widehat{\bm{\theta}}_{T})=0$. 

By allowing such constraint violation, we devise an efficient algorithm whose only projection is onto the hyper-plane  section $\bm{\Theta}_{1}^{d}$ of the feasible set $\bm{\Theta}^{d}$, and the constraint due to the size of the policy class $\bm{\Theta}_{R}^{d}$ is enforced implicitly using a Bregman divergence whose domain is subsumed by the set $\bm{\Theta}_{R}^{d}$. As we discuss in the next section, the Bregman projection onto the hyper-plane has an efficient implementation.

Notice, however, that due to the (potential) constraint violation of solutions $\widehat{\bm{\theta}}_{T}$, the vector $\bm{\Psi}\widehat{\bm{\theta}}_{T}$ may no longer be a valid occupancy measure. Nonetheless, it defines an admissible randomized policy via the following relation
\begin{align}
\label{Eq:Under_Policy}
\pi_{a}^{\widehat{\bm{\theta}}_{T}}(s)=\dfrac{[(\bm{\Psi}\widehat{\bm{\theta}}_{T})_{a}(s)]_{+} }{\sum_{a\in \mathcal{A}}[(\bm{\Psi}\widehat{\bm{\theta}}_{T})_{a}(s)]_{+} }, \quad \forall a\in \mathcal{A}, \forall s\in \mathcal{S}.
\end{align}
In the case that $(\bm{\Psi}\widehat{\bm{\theta}}_{T})_{a}(s)\leq 0$ for all pairs of action-state $(a,s)\in \mathcal{A}\times \mathcal{S}$, we define $\pi^{\widehat{\bm{\theta}}_{T}}_{a}(s)$ to be the uniform distribution. Let $\bm{\xi}^{\widehat{\bm{\theta}}_{T}}\df \bm{\xi}^{\bm{\pi}^{\widehat{\bm{\theta}}_{T}}}$ denotes the occupancy measure induced by the policy $\bm{\pi}^{\widehat{\bm{\theta}}_{T}}$ defined in Eq. \eqref{Eq:Under_Policy}, \textit{i.e.}, $\bm{\xi}^{\widehat{\bm{\theta}}_{T}}_{a}={\bm{\pi}_{a}^{\widehat{\bm{\theta}}_{T}}}\odot \bm{\mu}^{\bm{\pi}^{\widehat{\bm{\theta}}_{T}}}$.
\subsection{Expanded Efficiency} 

Equipped with the dual optimization problem in Eq. \eqref{Eq:dual_formulation},  we now describe the notion of efficiency of an algorithm to solve \eqref{Eq:dual_formulation}. The following definition is adapted from  \cite{abbasi2014linear}:

\begin{definition}\textsc{(Efficient Large Scale Dual ALP \cite{abbasi2014linear})}
	\label{Def:1}
	For an MDP specified by the cost matrix $\bm{c}$, probability transition matrix $\bm{P}$, a feature matrix $\bm{\Psi}$, the efficient large-scale dual ALP problem is to produce an estimate $\widehat{\bm{\theta}}\in \bm{\Theta}^{d}$ such that
	\begin{align}
	\label{Eq:A guarantee_for_that}
	\bm{c}^{T}\bm{\xi}^{\widehat{\bm{\theta}}} \leq \min_{\bm{\theta}\in \bm{\Theta}^{d}}\bm{c}^{T}\bm{\xi}^{\bm{\theta}}+\mathcal{O}(\varepsilon),
	\end{align} 
	in time polynomial in $d$ and $1/\varepsilon$ under the model of computation in (A.3). 
\end{definition} 

As described by Definition \ref{Def:1}, the computational complexity depends on the number of features $d$ only, and not the size of state-space $|\mathcal{S}|=n$.

The preceding definition is a special case of the following generalized definition that allows for the constraint violation:
\begin{definition}\textsc{(Expanded Efficient Large Scale Dual ALP \cite{abbasi2014linear})}
	\label{Def:2}
	Let $V:\real^{d}\mapsto \real_{+}$ be some violation function that measures how far $\bm{\Psi}\bm{\theta}$ is from a valid stationary distribution. In particular, $V(\bm{\theta})=0$ iff $\bm{\theta}\in \bm{\Theta}^{d}$ is a feasible point for the dual ALP in Eq. \eqref{Eq:Linear_Programming}. Then,
	\begin{align}
	\label{Eq:A guarantee_for_this} 
	\bm{c}^{T}\bm{\xi}^{\widehat{\bm{\theta}}}\leq \min_{\bm{\theta}\in \bm{\Theta}_{2}^{d}\cap \bm{\Theta}_{R}^{d}}\left[\bm{c}^{T}\bm{\xi}^{\bm{\theta}}+{1\over \varepsilon}V(\bm{\theta})\right] +\mathcal{O}(\varepsilon),
	\end{align} 
	in time polynomial in $d$ and $1/\varepsilon$, under the model of computation in (A.3).
\end{definition}
Clearly, a guarantee for \eqref{Eq:A guarantee_for_this} implies a guarantee for Eq. \eqref{Eq:A guarantee_for_that}. In addition, the expanded problem has a larger feasible set. Therefore, even if many feature vectors $\bm{\Psi}$ may not admit any feasible points in the feasible set $\bm{\Theta}^{d}$ and the dual problem is trivial, solving the expanded problem is still meaningful.

\subsection{Assumptions}
To establish our theoretical results, we need the following fast mixing assumption on underlying MDPs. This assumption implies that any policy quickly converges to its stationary distribution:
\vspace{1mm}      
\begin{itemize}
	\item[(\textbf{A.1})]
	\textsc{(Fast Mixing Markov Chain)}
	\label{Assumption:1}
	For $0<\varepsilon<1$, the Markov decision process specified by the tuple $M = (\mathcal{S},\mathcal{A},P, c,\gamma)$ is $\tau_{\text{mix}}$-mixing in the sense that	
	\small\begin{align}
	\label{Eq:Fast_Mixing}
	\hspace{-2mm}\tau_{\text{mix}}(\varepsilon)\df \max_{\bm{\pi}\in \Pi}\min\{t\geq 1: \| (\bm{P}^{\bm{\pi}})^{t}(s,\cdot)-\bm{\mu}_{\bm{\alpha}}^{\bm{\pi}}\|_{\text{TV}}\leq \varepsilon,\forall s\in \mathcal{S}\},
	\end{align}\normalsize
	for all $\pi\in \Pi$, where for two given Borel probability measures  $\bm{\nu},\bm{\mu}\in \mathcal{B}(A)$,  $\|\bm{\mu}-\bm{\nu}\|_{\text{TV}}\df \sup_{A\subset \mathcal{A}}|\bm{\mu}(A)-\bm{\nu}(A)|={1\over 2}\sum_{a\in A}|\mu(a)-\nu(a)|$ is the total variation norm.
\end{itemize}
The fast mixing condition \eqref{Eq:Fast_Mixing} is a standard assumption for Markov decision processes; see, \textit{e.g.}, \cite{abbasi2014linear}, \cite{wang2017primal}. The fast mixing condition \eqref{Eq:Fast_Mixing} implies that for any policy $\bm{\pi}\in \Pi^{d}$, there exists a constant $\tau_{\text{mix}}>0$ such that for all the distributions $\bm{\nu},\widehat{\bm{\nu}}$ over the state-action space,
\begin{align}
\|\bm{\nu}\bm{P}^{\bm{\pi}}-\widehat{\bm{\nu}}\bm{P}^{\bm{\pi}}\|_{\mathrm{TV}}\leq e^{-{1\over \tau_{\text{mix}}(\varepsilon)}}\|\bm{\nu}-\widehat{\bm{\nu}}\|_{\mathrm{TV}}. 
\end{align}

The following assumption is also standard in the literature; see, \textit{e.g.}, \cite{wang2017primal}:

\begin{itemize}
	\label{Assumption:2}
	\item[(\textbf{A.2})](\textsc{Uniformly Bounded Ergodicity}) The Markov decision process is ergodic under any stationary policy $\bm{\pi}$, and there exists $\kappa>0$ such that
	\begin{align}
	\label{Eq:Uniformly_Bounded}
	\dfrac{1}{n\sqrt{\kappa}} \bm{1} \leq \bm{\mu}_{\bm{\alpha}}^{\bm{\pi}}\leq \dfrac{\sqrt{\kappa}}{n}\bm{1}.
	\end{align}
	where we recall from Section \ref{Subsection:MDP} that $\bm{\mu}_{\bm{\alpha}}^{\bm{\pi}}$ is the stationary distribution over the state space of the MDP under the policy $\bm{\pi}$, and with the initial distribution $\bm{\alpha}$ on the states. 
\end{itemize}

Under the condition \eqref{Eq:Uniformly_Bounded} of (A.2), it is well-known that the underlying MDP is unichain \cite{puterman2014markov}, \textit{i.e.}, a Markov chain that contains a single recurrent class and possibly, some transient states.  Moreover, the factor $\kappa>0$ determines the amount of variation of stationary distributions associated with different policies, and thus can be sought of as a form of measuring the complexity of a MDP. Notice that in the case that some policies induce transient states (so the stationary distribution is not bounded away from zero), their mixture with an ergodic policy guarantee ergodicity.

Lastly, we consider the setup of the reinforcement learning in which the cost function $\bm{c}$ is unknown. But, the agent can interact with its environment and receives feedbacks from a sampling oracle:
\begin{itemize}
	\item[(\textbf{A.3})]\textsc{(Model-Free Reinforcement Learning)} We consider the following model of computation:
	\begin{enumerate}[leftmargin=*]
		\item[(\textit{i})] The state space $\mathcal{S}$, the action spaces $\mathcal{A}$, the reward upper bound and lower bounds $c_{\min}$ and $c_{\max}$, and the discount factor $\gamma$ are known.
		\item[(\textit{ii})] The cost function $\bm{c}$ is unknown.
		\item[(\textit{iii})] There is a sampling oracle that takes input $(a,s)\in \mathcal{A}\times \mathcal{S}$ and generates a new state $s'\in \mathcal{S}$ with probabilities $P_{a}(s',s)$ and returns the cost $c_{a}(s)\in [c_{\min},c_{\max}]$.
	\end{enumerate}

\end{itemize}

\section{Stochastic Primal-Dual Proximal Algorithm }
\label{Section:Main_Results}

In this section, we first describe a stochastic primal-dual proximal method to compute the coefficients of the mixture model in the dual optimization problem in Eq. \eqref{Eq:Dual_MixturePolicy}. We then analyze the efficiency as well as the sample complexity of the proposed algorithm.

\subsection{Stochastic Primal-Dual Algorithm}
\label{Subsection:MDP}

To apply the stochastic primal-dual method, we recast the optimization problem  \eqref{Eq:Dual_MixturePolicy} as the following MinMax optimization problem
\begin{align}
\label{Eq:Lagrangian}
\min_{\bm{\theta}\in \bm{\Theta}_{2}^{d}\cap \bm{\Theta}^{d}_{R}} \max_{\bm{\lambda}\in \real_{+}^{nm}} \mathcal{L}_{\bm{\Psi}}(\bm{\theta},\bm{\lambda})\df \bm{c}^{T}\bm{\Psi}\bm{\theta}-\bm{\lambda}^{T}\bm{\Psi}\bm{\theta}.
\end{align}
The following definition characterizes the notion of the saddle point of a MinMax problem:

\begin{definition}\ \textsc{(Saddle Point of MinMax Optimization Problem)}
Let $(\bm{\theta}^{\ast},\bm{\lambda}^{\ast})\in (\bm{\Theta}^{d}_{2}\cap \bm{\Theta}^{d}_{R})\times \real^{nm}_{+}$ denotes a \textit{saddle point} of the MinMax problem in Eq. \eqref{Eq:Lagrangian}, \textit{i.e.}, a point that satisfies the inequalities
\begin{align}
\label{Eq:MinMax}
\mathcal{L}_{\bm{\Psi}}(\bm{\theta}^{\ast},\bm{\lambda})\leq \mathcal{L}_{\bm{\Psi}}(\bm{\theta}^{\ast},\bm{\lambda}^{\ast})\leq \mathcal{L}_{\bm{\Psi}}(\bm{\theta},\bm{\lambda}^{\ast}),
\end{align}
for all $\bm{\theta}\in \bm{\Theta}^{d}_{2}\cap \bm{\Theta}^{d}_{R}$ and $\bm{\lambda}\in \real_{+}^{nm}$. 
\end{definition}

The primal optimal point $\bm{\theta}^{\ast}$ of the MinMax problem \eqref{Eq:Lagrangian} is an optimal solution for the dual problem \eqref{Eq:Dual_MixturePolicy}. 

Applying the stochastic primal-dual method to the MinMax problem \eqref{Eq:Lagrangian} is challenging as the the Lagrange multipliers $\bm{\lambda}\in \real_{+}^{nm}$ may take arbitrarily large values during the iterations of the primal-dual algorithm, resulting in a large sub-gradients for the Lagrangian function and instability in the performance. 
The following lemma, due to Chen and Wang  \cite[Lemma 1]{nedic2009approximate},  provides an upper bound on the norm of the optimal Lagrange multipliers:
\begin{lemma}\textsc{(Upper Bound on the Lagrange Multipliers, \cite[Lemma 1]{nedic2009approximate})}
	\label{Lemma:Wang}
Suppose $\bm{\lambda}^{\ast}\in \real^{nm}_{+}$ is a Lagrange multiplier vector satisfying the MinMax condition in Eq. \eqref{Eq:MinMax}.	Then, $\|\bm{\lambda}^{\ast}\|_{1}=\dfrac{\|\bm{\alpha}\|_{1}}{1-\gamma}=\dfrac{1}{1-\gamma}$, and $\sum_{a=1}^{m}\lambda_{a}(s)=p(s)$ for all $s\in \mathcal{S}$.
\end{lemma}

Now, define the norm ball $\bm{\Lambda}_{\beta}^{nm} \df \left\{\bm{\lambda}\in \real_{+}^{nm}:\|\bm{\lambda}\|_{2}\leq G_{\beta}\right\}$, where $G_{\beta}\df \dfrac{1}{1-\gamma}\vee \beta$. Here, $\beta>0$ is a free parameter of the algorithm that will be determined later. 

Lemma \ref{Lemma:Wang} suggests that the vector of optimal Lagrange multipliers belongs to the compact set $\bm{\Lambda}_{\beta}^{nm}$. Therefore, instead of the MinMax problem \eqref{Eq:Lagrangian}, we can consider the following equivalent problem
\begin{align}
\label{Eq:Lagrangian_1}
\min_{\bm{\theta}\in \bm{\Theta}_{2}\cap \bm{\Theta}_{R}} \max_{\bm{\lambda}\in \bm{\Lambda}} \mathcal{L}_{\bm{\Psi}}(\bm{\theta},\bm{\lambda}),
\end{align}
where the Lagrange multipliers $\bm{\lambda}$ are maximized over the compact set $\bm{\Lambda}_{\beta}^{nm}$ instead of entire non-negative orthant $\real_{+}^{nm}$. As we alluded earlier, the set of saddle points of the MinMax problems in Eqs. \eqref{Eq:Lagrangian_1} and \eqref{Eq:Lagrangian} coincide.

Algorithm \ref{Alg:1} describes a procedure to solve the MinMax problem in Eq. \eqref{Eq:Lagrangian_1}.  At each iteration $t\in [T]$ of the stochastic primal-dual method, we compute the following deterministic gradient 
\begin{align}
\label{Eq:Deterministic_Gradient}
\nabla_{\bm{\lambda}}\mathcal{L}_{\bm{\Psi}}(\bm{\theta}_{t},\bm{\lambda}_{t})=\bm{\Psi}\bm{\theta}_{t}.
\end{align}
Furthermore,  we draw a random index $i\sim {\textsc{Uniform}}\{0,1,2,\cdots,d\}$, and subsequently sample the state-action $(s_{t},a_{t})\sim \bm{\psi}^{i}$.  Then, we compute the stochastic gradient as follows
\begin{subequations}
	\label{Eq:Stochastic_Gradient}
	\small\begin{align}
	\label{Eq:Stochastic_Gradient_1}
	\nabla_{\bm{\theta}} \widehat{\mathcal{L}}_{\bm{\Psi}}(\bm{\theta}_{t},\bm{\lambda}_{t})&=\dfrac{c_{a_{t}}(s_{t})\bm{\Psi}_{a_{t}}(s_{t})-\lambda_{t,a_{t}}(s_{t})\bm{\Psi}_{a_{t}}(s_{t})}{{1\over d}\sum_{i=1}^{d}\psi_{a_{t}}^{i}(s_{t})},
	\end{align}\normalsize
\end{subequations} 
where $\bm{\Psi}_{a_{t}}(s_{t})\in \real^{d}$ is a row vector, corresponding to the row $(s_{t},a_{t})$ of the feature matrix $\bm{\Psi}\in \real^{nm\times d}$. We notice that $\nabla_{\bm{\theta}} \widehat{\mathcal{L}}_{\bm{\Psi}}(\bm{\theta}_{t},\bm{\lambda}_{t})$ is an unbiased estimator of the gradients of the Lagrangian function $\nabla_{\bm{\theta}}\mathcal{L}_{\bm{\Psi}}(\bm{\theta},\bm{\lambda})$. Formally,
\begin{align}
\nonumber\expect\left[\nabla_{\bm{\theta}} \widehat{\mathcal{L}}_{\bm{\Psi}}(\bm{\theta}_{t},\bm{\lambda}_{t})\right]&=\hspace{-4mm}\sum_{(s,a)\in \mathcal{S}\times \mathcal{A}}\hspace{-4mm}\dfrac{c_{a_{t}}(s_{t})\bm{\Psi}_{a_{t}}(s_{t})-\lambda_{t,a_{t}}(s_{t})\bm{\Psi}_{a_{t}}(s_{t})}{{1\over d}\sum_{i=1}^{d}\psi_{a_{t}}^{i}(s_{t})}\prob[s_{t}=s,a_{t}=a]\\  \nonumber
&=\hspace{-4mm}\sum_{(s,a)\in \mathcal{S}\times \mathcal{A}}\hspace{-4mm}{c_{a}(s)\bm{\Psi}_{a}(s)-\lambda_{t,a}(s)\bm{\Psi}_{a}(s)}\\ \label{Eq:Unbiased_Estimator}
&=\bm{c}^{T}\bm{\Psi}-\bm{\lambda}_{t}^{T}\bm{\Psi}= \nabla_{\bm{\theta}}\mathcal{L}_{\bm{\Psi}}(\bm{\theta}_{t},\bm{\lambda}_{t}).
\end{align}
Algorithm \ref{Alg:1} describes the primal-dual steps to update $(\bm{\theta}_{t},\bm{\lambda}_{t})$. To implicitly control the size of the policy class as characterized by the constraint $\bm{\Theta}_{R}^{d}$, in Algorithm \ref{Alg:1} we consider a Bregman divergence whose modulus has the domain $\text{dom}(\phi)=\bm{\Theta}_{R}^{d}\cup \{+\infty\}$. For example, when the policy class is defined by a Euclidean ball $\bm{\Theta}_{R}^{d}\df \{\bm{\theta}\in \real^{d}:\|\bm{\theta}\|_{2}\leq R\}$, the following convex function can be employed (see \cite{dhillon2007matrix})
\begin{align}
\label{Eq:modulus}
\phi(\bm{\theta})=-\sqrt{R^{2}-\|\bm{\theta}\|_{2}^{2}}.
\end{align}
The associated convex conjugate of the modulus \eqref{Eq:modulus} is
\begin{align}
\label{Eq:Convex_Conjugate_Of_Hellinger}
\phi^{\ast}(\bm{\theta})=R\sqrt{1+\|\bm{\theta}\|_{2}^{2}}.
\end{align}
Moreover, the modulus \eqref{Eq:modulus} yields the Hellinger-like divergence
\begin{align}
\label{Eq:Hellinger}
D_{\phi}(\bm{\theta}_{1},\bm{\theta}_{2})= \dfrac{R^{2}-\langle \bm{\theta}_{1},\bm{\theta}_{2} \rangle}{\sqrt{R^{2}-\|\bm{\theta}_{2}\|_{2}^{2}}} -\sqrt{R^{2}-\|\bm{\theta}_{1}\|_{2}^{2}}.
\end{align}
Alternatively, when the policy class is defined by a hyper-cube $\bm{\Theta}^{d}_{R}\df \{\bm{\theta}\in \real^{d}:-R\leq \theta_{i}\leq R,i=1,2,\cdots,d\}$, a function similar to the symmetric logistic function yields the divergence with the desired domain. In particular, let
\begin{align}
\label{Eq:bit_entropy}
\hspace{-1mm}\phi(\bm{\theta})=\sum_{i=1}^{d}\Big[(R+\theta_{i})\log(R+\theta_{i})+(R-\theta_{i})\log(R-\theta_{i})\Big].
\end{align}
The convex conjuagte of $\phi(\bm{\theta})$ is the following function
\begin{align}
\phi^{\ast}(\bm{\theta})=-\sum_{i=1}^{d}R\theta_{i} -\sum_{i=1}^{d}2R\log\left(\dfrac{2R}{e^{\theta_{i}}+1}\right).
\end{align}
The Bregman divergence associated with the modulus in Eq. \eqref{Eq:bit_entropy} is
\begin{align}
D_{\phi}(\bm{\theta}_{1},\bm{\theta}_{2})=\sum_{i=1}^{d}\left[(\theta_{1}^{i}+R)\log\dfrac{(\theta_{1}^{i}+R)}{(\theta_{2}^{i}+R)} +(R-\theta_{1}^{i})\log\dfrac{(R-\theta_{1}^{i})}{(R-\theta_{2}^{i})}\right].
\end{align}
The steps \eqref{Eq:Recursion_1}-\eqref{Eq:Recursion_3} of Algorithm \ref{Alg:1} are reminiscent of the so-called \textsc{Mirror Descent Algorithm} for online convex optimization \cite{beck2003mirror},
\begin{align}
\label{Eq:Compact_Update_Rule}
\hspace{-2mm}{\bm{\theta}_{t+1}}=\arg\min_{\bm{\theta}\in \bm{\Theta}_{2}} \left\{\langle \nabla_{\bm{\theta}}\mathcal{L}_{\bm{\Psi}}(\bm{\theta}_{t},\bm{\lambda}_{t}),\bm{\theta}-\bm{\theta}_{t} \rangle+\dfrac{1}{\eta_{t}}D_{\phi}(\bm{\theta};\bm{\theta}_{t})\right\},
\end{align}
However, the role of the Bregman divergence in Algorithm \ref{Alg:1} is different from that of the \textsc{Mirror Descent} algorithm. In the \textsc{Mirror Descent} algorithm, the Bregman divergence is typically used to adapt to the geometry of the feasible set and achieve better dependency on the dimension of the embedding space of the feasible set. In contrast, the Bregman divergence  in Algorithm \ref{Alg:1} is employed to implicitly enforce the constraints due to the size of the policy class and eliminate the projection step.

To see this equivalence, first notice that the update rule in Eq. \eqref{Eq:Compact_Update_Rule} can be rewritten as below
\begin{subequations}
	\begin{align}
	\label{Eq:Bregman_Projection_H0}
	\widetilde{\bm{\theta}}_{t}&=\nabla\phi^{\ast}\big(\nabla\phi(\bm{\theta}_{t})-\eta_{t}\nabla_{\bm{\theta}}\widehat{\mathcal{L}}_{\bm{\Psi}}(\bm{\theta}_{t},\bm{\lambda}_{t})\big),\\ \label{Eq:Bregman_Projection_H}
	\bm{\theta}_{t+1}&=\arg\min_{\bm{\theta}\in \bm{\Theta}_{2}}D_{\phi}(\bm{\theta};\widetilde{\bm{\theta}}_{t}),
	\end{align}
\end{subequations}
where we also recall $\bm{\Theta}_{2}^{d}=\{\bm{\theta}\in \real^{d}:\bm{\theta}^{T}\cdot\bm{1}=1\}$. Now, consider the Bregman projection  \eqref{Eq:Bregman_Projection_H} onto the hyper-plane $\bm{\Theta}_{2}^{d}$. As shown in \cite{dhillon2007matrix}, the Bregman projection in Equation \eqref{Eq:Bregman_Projection_H} can be alternatively computed using Eqs. \eqref{Eq:Recursion_2}-\eqref{Eq:Recursion_3}. To demonstrate this, first invoke the necessary and sufficient KKT conditions for the optimality of $\bm{\theta}_{t+1}$ which requires the existence of an optimal Lagrange multiplier $z_{t}\in \real$ that satisfies the following equation
\begin{align}
\label{Eq:KKT}
\nabla_{\bm{\theta}}D_{\phi}(\bm{\theta};\widetilde{\bm{\theta}}_{t})\vert_{\bm{\theta}=\bm{\theta}_{t+1}}=z_{t} \nabla_{\bm{\theta}} (\bm{\theta}^{T}\bm{1}-1).
\end{align}
From Eq. \eqref{Eq:KKT}, we establish that
\begin{align}
\nabla\phi(\bm{\theta}_{t+1})=z_{t}\bm{1}+\nabla\phi(\widetilde{\bm{\theta}}_{t}).
\end{align}
Alternatively, since the gradient of a Legendre function $\phi$ is a bijection from $\text{dom}(\phi)$ to $\text{dom}(\phi^{\ast})$ and its inverse is the gradient of the conjugate (cf. \cite[Thm. 26.5]{rockafellar2015convex}), we have 
\begin{align}
\label{Eq:FROMFROM}
\bm{\theta}_{t+1}=\nabla\phi^{\ast}\left(z_{t}\bm{1}+\nabla\phi(\widetilde{\bm{\theta}_{t}}) \right).
\end{align}
Since $\bm{\theta}_{t+1}^{T}\cdot\bm{1}=1$, from Eq. \eqref{Eq:FROMFROM} we obtain
\begin{align}
\nabla\phi^{\ast}\left(z_{t}\bm{1}+\nabla\phi(\widetilde{\bm{\theta}_{t}}) \right)^{T}\cdot \bm{1}-1=0.
\end{align}
The left hand side is the derivative of the strictly convex function $J(z)$ defined below
\begin{align}
\label{Eq:minimization}
J(z)\df  \phi^{\ast}\left(z\bm{1}+\nabla\phi(\widetilde{\bm{\theta}}_{t}) \right)-z
\end{align}
Therefore, Equation \eqref{Eq:FROMFROM} implies that 
\begin{align}
z_{t}=\arg\min_{z\in \real} J(z).
\end{align}
Furthermore, $z_{t}$ is the unique minimizer of $J(z)$.

In some special cases, the solution of the minimization problem in Eq. \eqref{Eq:minimization} can be computed explicitly. For instance, using the modulus in Eq. \eqref{Eq:modulus} and its associated convex conjugate function in Eq. \eqref{Eq:Convex_Conjugate_Of_Hellinger}, we derive that
\begin{align}
J(z)=R\left(1+\left\|z\bm{1}+\dfrac{\widetilde{\bm{\theta}}_{t}}{\sqrt{R^{2}-\|\widetilde{\bm{\theta}}_{t}\|_{2}^{2}} }  \right\|_{2}^{2}\right)^{1\over 2}-z.
\end{align}
Hence, the minimization in Eq. \eqref{Eq:minimization} has the following explicit solution 
\small\begin{align}
\label{Eq:zt}
&z_{t}=\dfrac{1}{(Rd)^{2}-d}\Bigg(-\bm{1}^{T}\bm{V}_{t}\cdot (R^{2}d-1)
+\sqrt{(\bm{1}^{T}\bm{V}_{t}\cdot (R^{2}d-1))^{2}-((Rd)^{2}-d)(R^{2}(\bm{1}^{T}\bm{V}_{t}-\|\bm{V}_{t}\|_{2}^{2}-1) ) } \Bigg),
\end{align}\normalsize
where $\bm{V}_{t}=\widetilde{\bm{\theta}}_{t}/\sqrt{R^{2}-\|\widetilde{\bm{\theta}}_{t}\|_{2}^{2}}$. Therefore, Algorithm \ref{Alg:1}  is projection-free and for the modulus given in Eq. \eqref{Eq:modulus}, the minimization step in \eqref{Eq:Recursion_2} of Algorithm \ref{Alg:1} is redundant. As a result, each iteration of Algorithm \ref{Alg:1} can be computed very efficiently and cheaply.

Notice that when $R=1/\sqrt{d}$, the function $J(z)$ is monotonic decreasing, and $J(z)> 0$ and $\lim_{z\rightarrow \infty}J(z)=0$.

For a general modulus $\phi$, the following properties hold for $J(z)$ (see  \cite{dhillon2007matrix}):
\begin{itemize}
	\item[(P.1)] The domain of $J$ contains a neighborhood of zero since $J(0)=\langle \bm{y},\nabla\phi(\bm{y}) \rangle-\phi(\bm{y})$.
	\item[(P.2)] Since $\phi^{\ast}$ is a Legendre function, the first derivative of $J$ always exists, and is given by
	\begin{align}
	J'(z)=\nabla\phi^{\ast}(z\bm{1}+\nabla\phi(\bm{y}))^{T}\bm{1}-1.
	\end{align}
	\item[(P.3)] When the Hessian of $\phi^{\ast}$ exists, we 
	\begin{align}
	J''(z)=\bm{1}^{T}\nabla^{2}\phi^{\ast}\Big(z\bm{1}+\nabla\phi(\widetilde{\bm{\theta}}_{t})\Big)\bm{1}.
	\end{align}
	\item[(P.4)] When the modulus $\phi$ is separable (as in Eq. \eqref{Eq:bit_entropy}), the Hessian matrix $\nabla^{2}\phi^{\ast}$ is diagonal. 
\end{itemize}

\begin{algorithm}[t!]\footnotesize{
	\footnotesize  \caption{\small{Projection-Free Primal-Dual Proximal Method for the Dual ALP}}
	\begin{algorithmic}
		\State {\bfseries Inputs:} Number of rounds $T\in \mathbb{N}$ and the size of the policy class $R>0$. A sequence of non-increasing, non-negative values  $\{\eta_{t}\}_{t=1}^{T}$ for the learning rate.
	
		\State {\bfseries Initialize:} Initial values $\bm{\theta}_{0}\in\bm{\Theta}_{2}^{d}\cap \bm{\Theta}^{d}_{R}$ and $\bm{\lambda}_{0}=\bm{0}\in \bm{\Lambda}_{\beta}^{nm}$ for a chosen $\beta>0$. 
		\For{$t=1,2,\cdots,T$}
		\State{(1)-Sample a base policy $i\in [d]$ uniformly, and let $(s_{t},a_{t})\sim \bm{\psi}^{i}$}
		\State{(2)-Incur the cost $c_{a_{t}}(s_{t})$.}
		\State{(3)-Compute the gradients $\nabla_{\bm{\lambda}} \mathcal{L}_{\bm{\Psi}}(\bm{\theta}_{t},\bm{\lambda}_{t})$ and $\nabla_{\bm{\theta}}\widehat{\mathcal{L}}_{\bm{\Psi}} (\bm{\theta}_{t},\bm{\lambda}_{t})$ from Eqs. \eqref{Eq:Deterministic_Gradient} and \eqref{Eq:Stochastic_Gradient_1}. }
		\State{(4)-Update the primal vector:}
		\begin{subequations}
			\label{Eq:Projection}
			\footnotesize  \begin{align}
			\label{Eq:Recursion_1}
			\widetilde{\bm{\theta}}_{t}&=\nabla\phi^{\ast}\left(\nabla \phi(\bm{\theta}_{t})-\eta_{t}\nabla_{\bm{\theta}}\widehat{\mathcal{L}}_{\bm{\Psi}}(\bm{\theta}_{t},\bm{\lambda}_{t})\right).\\ \label{Eq:Recursion_2}
			z_{t}&=\arg\min_{z\in \real} J(z)= \phi^{\ast}\left(z\bm{1}+\nabla\phi(\widetilde{\bm{\theta}}_{t})\right)-z.\\ 	\label{Eq:Recursion_3}
			\bm{\theta}_{t+1}&=\nabla\phi^{\ast}\left(z_{t}\bm{1}+\nabla\phi(\widetilde{\bm{\theta}}_{t})\right).
			\end{align}
		\end{subequations} 
		\State{(5)-Update the Lagrange multipliers vector:}
		\begin{align}
		\label{Eq:Update_Rule_For_Lagrange_Multipliers}
		\bm{\lambda}_{t+1}=\dfrac{G_{\beta}\cdot\Big(\bm{\lambda}_{t}+\eta_{t}\nabla_{\bm{\lambda}}\widehat{\mathcal{L}}_{\bm{\Psi}}(\bm{\theta}_{t},\bm{\lambda}_{t}) \Big)}{G_{\beta}\vee\|\bm{\lambda}_{t}+\eta_{t}\nabla_{\bm{\lambda}}\widehat{\mathcal{L}}_{\bm{\Psi}}(\bm{\theta}_{t},\bm{\lambda}_{t})\|_{2}}, \quad \widehat{G}_{\beta}\df \dfrac{1}{1-\gamma}\vee \beta.
		\end{align}
		\State{(6)-Compute the running averages:}
		\begin{subequations}
			\label{Eq:Running_averages}
			\begin{align}
			\widehat{\bm{\theta}}_{T}= \dfrac{\sum_{t=0}^{T-1}\eta_{t}\bm{\theta}_{t} }{\sum_{t=0}^{T-1}\eta_{t} }, \quad 
			\widehat{\bm{\lambda}}_{T}=\dfrac{\sum_{t=0}^{T-1}\eta_{t} \bm{\lambda}_{t}}{\sum_{t=0}^{T-1}\eta_{t}}.			 
			\end{align}
		\end{subequations}
		\EndFor   
		\State{\textbf{Output}}: The randomized policy:
		\begin{align}
		\label{Eq:Alg_R_P}
		\pi_{a}^{\widehat{\bm{\theta}}_{T}}(s)=\dfrac{[(\bm{\Psi}\widehat{\bm{\theta}}_{T})_{a}(s)]_{+} }{\sum_{a\in \mathcal{A}}[(\bm{\Psi}\widehat{\bm{\theta}}_{T})_{a}(s)]_{+} }, \quad \forall a\in \mathcal{A}, \forall s\in \mathcal{S}\normalsize.
		\end{align}	
		
	\end{algorithmic}
	\label{Alg:1}}
\end{algorithm}\normalsize

\subsection{Analysis of Efficiency}

We now analyze the efficiency of the stochastic primal-dual algorithm in Algorithm \ref{Alg:1}. We postpone the proofs of the following results to Section \ref{Section:Proof_of_Main_Results}. 

To state the first lemma, we need a few definitions. We define the diameter of the policy class as
\begin{align}
D\df \max_{\bm{\theta}_{1},\bm{\theta}_{2}\in \bm{\Theta}_{R}^{d}}D(\bm{\theta}_{1};\bm{\theta}_{2}).
\end{align}
Furthermore, we define the following constant that appears in the convergence bound
\begin{align}
\label{Constant:A}
A\df \max_{(s,a)\in \mathcal{S}\times \mathcal{A}} \left\|\dfrac{\bm{\Psi}_{a}(s)}{{1\over d}\sum_{i=1}^{d}\psi_{a}^{i}(s)}\right\|_{\ast},
\end{align}
where $\|\cdot\|_{\ast}$ is the dual norm associated with the norm $\|\cdot\|$ characterizing the policy class $\bm{\Theta}_{R}^{d}$.

Now, we are in position to state the first lemma of this paper:
\begin{proposition}\textsc{(High Probability and in Expectation Convergence)}
	\label{Proposition:Convergence_Alg1}
	Suppose conditions $\mathbf{(A.1)}$, $\mathbf{(A.2)}$, and $\mathbf{(A.3)}$ are satisfied. Consider $T$ rounds of Algorithm \ref{Alg:1}, and let $\widehat{\bm{\theta}}_{T}$ denotes the running average in Eq. \eqref{Eq:Running_averages}. Recall that $G_{\beta}=\dfrac{1}{1-\gamma}\vee \beta$, where $\beta>0$ is a free parameter.
	\begin{itemize}
		\item[(\textit{i})] With the probability of at least $1-{4\over T}$, the following inequality holds for $T\geq 5$,
		\small\begin{align}
\label{Eq:Bound_1}
\bm{c}^{T}\bm{\Psi}\widehat{\bm{\theta}}_{T}- \bm{c}^{T}\bm{\Psi}\bm{\theta}^{\ast}\leq \dfrac{1}{2\sqrt{T}-1} \Bigg(&D+4B\ln(T)+4\sqrt{2}{RC_{e}\over c_{0}}\ln(T) 	\\ \nonumber
&+{C_{e}C(n)}\max\left\{\sqrt{2\over c_{1}}\ln(T),{4\over c_{1}}\ln^{2}(T) \right\}\Bigg)\df \alpha_{T}.
		\end{align}\normalsize
		\item[(\textit{ii})] In the expectation, the following inequality holds for $T\geq 2$,
		\small\begin{align}
		\label{Eq:Bound_2}
		\hspace{-1 mm}\expect\Big[\bm{c}^{T}\bm{\Psi}\widehat{\bm{\theta}}_{T}\Big]- \bm{c}^{T}\bm{\Psi}\bm{\theta}^{\ast}\leq \dfrac{D+B(1+\ln(T))}{2\sqrt{T}-1},              
		\end{align}\normalsize
		where the expectation is taken with respect to the randomness.
		\item[\textit{(iii)}]  Define the constraint violation function $V(\widehat{\bm{\theta}}_{T})\df {1\over 2} \big\|\big[\bm{\Psi}\widehat{\bm{\theta}}_{T}\big]_{-}\big\|_{2}$. Then, for the choice of the hyper-parameter $\beta>(1-\gamma)^{-1}$, we have
		\small{\begin{align}
			V(\widehat{\bm{\theta}}_{T})\leq {Rc_{\max}\sqrt{nm}\|\bm{\Psi}\|_{\ast,2}\over \beta}+{1\over 2\beta}\alpha_{T}+\dfrac{\beta}{2(\sqrt{T}-1) }.		\label{Eq:Constraint_Violation_Bound}
			\end{align}}
		
 \end{itemize}
The constants $B$ and $C_{e}$ in the upper bounds \eqref{Eq:Bound_1}-\eqref{Eq:Bound_2}, are defined as follows
	\begin{subequations}
		\begin{align}
		\label{Eq:Constant_C}
		B&\df 2R\sigma_{\max}(\bm{\Psi})+(\sqrt{nm}+G_{\beta})\|\bm{\Psi}\|_{\ast,2}+C_{e}^{2},\\
		C_{e}&\df (1+G_{\beta})A+(\sqrt{nm}+G_{\beta})\|\bm{\Psi}\|_{\ast,2},
		\end{align}
		respectively. Moreover, $c_{0},c_{1}>0$ are some universal constants, and $C(n)$ is a constant depending on $n$, such that $\|\bm{v}\|\leq C(n)\|\bm{v}\|_{2}$ for all $\bm{v}\in \real^{n}$.
	\end{subequations}
	\hfill $\square$
\end{proposition}

 In Proposition \ref{Proposition:Convergence_Alg1}, we established the rate of convergence to the optimal objective value of the dual optimization problem in Eq.  By choosing the free parameter as $\beta=\mathcal{O}(T^{1\over 4})$, then 

 and thus $V(\widehat{\bm{\theta}}_{T})\rightarrow 0$ as $T\rightarrow \infty$. In addition,  with the same choice of $\beta=\mathcal{O}(T^{1\over 4})$, the following convergence rate is attained
 \begin{align}
 \bm{c}^{T}\bm{\Psi}\widehat{\bm{\theta}}_{T}- \bm{c}^{T}\bm{\Psi}\bm{\theta}^{\ast}=\mathcal{O}\left(\dfrac{\ln(T)}{T^{1\over 4}}\right).
 \end{align}

The next lemma due to \cite{abbasi2014linear} relates the amount of constraint violation of the vector $\bm{\Psi}\widehat{\bm{\theta}}_{T}$ to its distance from the occupation measure $\bm{\xi}^{{\widehat{\bm{\theta}}}_{T}}$:
\begin{lemma}\ \textsc{(Distance from the Occupation Measure),  \cite[Lemma 2]{abbasi2014linear}}
	\label{Lemma:Abbasi}
	Suppose conditions $\mathbf{(A.1)}$, $\mathbf{(A.2)}$ and $\mathbf{(A.3)}$ are satisfied. Let $\bm{\Psi}\widehat{\bm{\theta}}_{T}\in \real^{nm}$ and $\bm{\pi}^{\widehat{\bm{\theta}}_{T}}$ be the outputs of Algorithm \ref{Alg:1}. Further, let $\bm{\xi}^{{\widehat{\bm{\theta}}}_{T}}$ denotes the occupation measure in conjunction with the policy $\pi_{a}^{\widehat{\bm{\theta}}_{T}}(s)$. Then, the following inequality hold
	\begin{align}
	\label{Eq:Distance_From_The_Occupation_Measure}
	\hspace{-3mm}	\|\bm{\Psi}\widehat{\bm{\theta}}_{T}- \bm{\xi}^{\widehat{\bm{\theta}}_{T}}\|_{\mathrm{TV}}\leq \dfrac{2\tau_{\mix}V(\widehat{\bm{\theta}}_{T}) \log(1/V(\widehat{\bm{\theta}}_{T}))+3V(\widehat{\bm{\theta}}_{T})}{1-\gamma},
	\end{align}	
	where we recall that $V(\bm{\theta})={1\over 4}\big\| [\bm{\Psi}\widehat{\bm{\theta}}_{T}]_{-} \big\|_{2}$ is the constraint violation.	
	$\hfill$	$\square$
\end{lemma}

Equipped with Proposition \ref{Proposition:Convergence_Alg1} and Lemma \ref{Lemma:Abbasi}, we are in position to state the main theorem of this paper. Recall the notion of efficiency and expanded efficiency from Definitions \ref{Def:1} and \ref{Def:2}, respectively. The following theorem establishes a probability approximately correct lower bound on the number of rounds required for Algorithm \ref{Alg:1} to achieve near optimal objective value of the dual optimization problem:

\begin{theorem}\ \textsc{(PAC Sample Complexity for $\varepsilon$-Optimal Efficiency)}
	\label{Thm:Main_Theorem}	
	Suppose the conditions $\mathbf{(A.1)}$, $\mathbf{(A.2)}$, and $\mathbf{(A.3)}$ are satisfied. Consider $T$ rounds of Algorithm \ref{Alg:1}, where
	\begin{align}
	T=\Omega\left(\dfrac{\tau_{\mix}^{4}nm \|\bm{\Psi}\|_{\ast,2}^{4} }{(1-\gamma)^{4}\varepsilon^{4}}\right).
	\end{align}	
	Then, the stationary measure $\bm{\xi}^{\widehat{\bm{\theta}}_{T}}$ associated with the randomized policy $\bm{\pi}^{\widehat{\bm{\theta}}_{T}}$ generated by Algorithm \ref{Alg:1} satisfies the following expanded efficiency inequality 
	\begin{align}
	\bm{c}^{T}\bm{\xi}^{\widehat{\bm{\theta}}_{T}}\leq \bm{c}^{T}\bm{\xi}^{\ast}+\varepsilon,
	\end{align}
	with the probability of at least $1-\rho$.
	\hfill $\square$
\end{theorem}

The proof of Theorem \ref{Thm:Main_Theorem} follows directly from Proposition \ref{Proposition:Convergence_Alg1} and \ref{Lemma:Abbasi}. We notice that the geometry of the policy class influences the sample complexity through the subordinate norm $\|\bm{\Psi}\|_{\ast,2}$.

\section{Connection to Prior Works}

Earlier works on the approximate linear programming have focused on the discounted version of the primal problem associated with the Bellman equation. Let $\mathscr{S}:\real^{n}\rightarrow \real^{n}$ denotes the Bellman operator defined by its action $(\mathscr{S}v)(s) =\min_{a\in \mathcal{A}}\{c_{a}(s)+\gamma \sum_{s\in \mathcal{S}}P_{a}(s,s')v(s')\}$ for $s\in \mathcal{S}$. Let $\bm{\Psi}\in \real^{n\times d}$ be a feature matrix. de Farias and Van Roy \cite{de2004constraint} study the discounted version of the primal problem
\begin{subequations}
	\label{Eq:Exact_Inexact}
\begin{align}
&(\text{Exact Primal LP}):\min_{\bm{v}\in \real^{d}} \bm{p}^{T}\bm{v}   \hspace{5mm}    (\text{Approximate Primal LP}): \min_{\bm{\nu}\in \real^{d}} \bm{p}^{T}\bm{\Psi}\bm{\nu}\\ \label{Eq:Large_Number_Of_Constraints}
&\hspace{24.3mm}\text{s.t.}:  \mathscr{S}\bm{v}\succeq \bm{v}.  \hspace{41mm}   \text{s.t.}:  \mathscr{S}\bm{\Psi}\bm{\nu}\succeq \bm{\bm{\Psi}}\bm{\nu}.
\end{align}
\end{subequations}
respectively. Notice that upon expansion of $\mathscr{S}\bm{v}$, the exact LP in Eq. \eqref{Eq:Exact_Inexact} has the same form as the optimization in Eq. \eqref{Eq:primal_problem}.

Recall the definition of the value function $v^{\bm{\pi}}(s)=\lim_{T\rightarrow \infty}\expect\big[\sum_{t=0}^{T}\gamma^{t}c_{\pi(s_{t})}(s_{t})\big|s_{0}=s\big]$. de Farias and Van Roy \cite[Thm. 1]{de2004constraint} show that for any vector $\bm{v}\in \real^{n}$ satisfying the constraint $\mathscr{S}\bm{v}\succeq \bm{c}$, we have 
\begin{align}
\expect_{s\sim \bm{\alpha}}\big[v^{\bm{\pi}}(s)-v^{\ast}(s)\big] \leq \dfrac{1}{1-\gamma}\sum_{s\in \mathcal{S}}\mu^{\bm{\pi}}(s)|v^{\bm{\pi}}(s)-v^{\ast}(s)|.
\end{align}
where the expectation on the left hand side is taken with respect to the initial distribution $\bm{\alpha}\in \Delta_{n}$ over the states $\mathcal{S}$.

Define the \textit{Lyapunov function} $\bm{u}:\mathcal{S}\rightarrow \real$, and $\beta_{\bm{u}}\df \gamma \max_{(s,a)\in \mathcal{S}\times \mathcal{A}} {\sum_{s\in \mathcal{S}}P_{a}(s,s')u(s)\over u(s)}$. Let $\bm{\nu}^{\ast}\in \real^{n}$ denotes the solution of the approximate LP. Then, for any Lyapunov function $\bm{u}\in \mathcal{U}_{\bm{\Psi}}\df \{\bm{u}\in \real^{n}:\bm{u}\succeq \bm{1},\bm{u}\in \mathrm{Span}(\bm{\Psi}),\beta_{\bm{u}}<1\}$, de Farias and Van Roy \cite[Thm. 3]{de2004constraint} have shown that
\begin{align}
\label{Eq:Tighter_Than_This}
\expect_{s\sim \bm{\alpha}}[|v^{\ast}(s)-(\bm{\Psi}\bm{\nu}^{\ast})(s)|]\leq \inf_{\bm{u}\in \mathcal{U}_{\bm{\Psi}}} \dfrac{2\bm{\alpha}^{T}\bm{u}}{1-\beta_{\bm{u}}} \min_{\bm{\nu}\in \Gamma}\max_{s\in \mathcal{S}}\dfrac{1}{(\bm{\Psi}\bm{u})(s)} |v^{\ast}(s)-(\bm{\Psi}\bm{\nu})(s)|. 
\end{align}

For large action/state spaces, the number of constraints in Eq. \eqref{Eq:Large_Number_Of_Constraints} is large and solving ALP can be computationally prohibitive. To address this issue, de Farias and Van Roy \cite{de2004constraint} propose an efficient algorithm based on the constraint sampling technique (cf. Thm. \ref{Thm:de_Farias}). For some Lyapunov function $\bm{u}\in \mathcal{U}_{\bm{\Psi}}$, consider the following constraint sampling distribution over states-actions, 
\begin{align}
\eta(s,a)\df \dfrac{\mu^{\pi^{\ast}}_{\bm{\alpha}}(s)u(s)}{m\cdot\bm{\mu}^{\pi^{\ast}}_{\bm{\alpha}}\bm{u}^{T}},\quad (s,a)\in \mathcal{S}\times \mathcal{A}.
\end{align}
Moreover, define
\begin{align}
\label{Eq:Constraint_Sampling_Scheme}
\zeta\df \dfrac{1+\beta_{\bm{u}}}{2}\dfrac{(\bm{\mu}^{\bm{\pi}^{\ast}}_{\bm{\alpha}})^{T}\bm{u}  }{\bm{\alpha}^{T}\bm{v}^{\ast}} \sup_{\bm{\nu}\in \Gamma} \dfrac{1}{u(s)} |v^{\ast}(s)-\bm{\Psi}\bm{\nu}| . 
\end{align}
where $\Gamma\subset \real^{d}$. Consider 
\begin{align}
n_{0}=\Omega\left( \dfrac{16m\zeta}{(1-\gamma)\epsilon}\left(d\log\left(\dfrac{48m\zeta}{(1-\gamma)\epsilon} \right) \right)+\log\dfrac{2}{\delta}\right),
\end{align}
random state-action pairs sampled under the distribution $\eta$, and let $\mathcal{N}\subseteq \mathcal{S}\times \mathcal{N}$ denotes the corresponding set.  de Farias and Van Roy \cite{de2004constraint} propose to solve the following optimization problem
\begin{align}
&\min_{\bm{\nu}\in \Gamma} \bm{\alpha}^{T}\bm{\Psi}\bm{\nu}\\
&\text{s.t.}: c_{a}(s)\geq (\bm{\Psi}\bm{\nu})(s)-\gamma\bm{P}_{a}(s)\bm{\Psi}\bm{\nu}, \quad (a,s)\in \mathcal{N},
\end{align}
Then, it is shown in \cite{de2004constraint} that with the probability of at least $1-\delta$, the following inequality holds
\begin{align}
\expect_{s\sim \bm{\alpha}}[|v^{\ast}(s)-(\bm{\Psi}\widehat{\bm{\nu}}^{\ast})(s)|]\leq \expect_{s\sim \bm{\alpha}}[|v^{\ast}(s)-(\bm{\Psi}\widehat{\bm{\nu}}^{\ast})(s)|]+\epsilon \expect_{s\sim \bm{\alpha}}[v^{\ast}(s)].
\end{align}
Notice, however, that the constraint sampling distribution in Eq. \eqref{Eq:Constraint_Sampling_Scheme} depends on the stationary distribution $\mu_{\bm{\alpha}}^{\bm{\pi}^{\ast}}$ associated with the optimal policy $\bm{\pi}^{\ast}$. However, the optimal policy is unknown a priori, rendering the sampling scheme in \cite{de2004constraint} restrictive for many real applications.

Desai \textit{et al.} \cite{desai2012approximate} studied a smoothed approximate linear program, in which slack variables are added to the linear constraints that allow for their violations. In particular, the smoothed ALP (SALP) has the following form 
\begin{align}
&\min_{(\bm{\nu},\bm{s})\in \Gamma\times \real^{n}} \bm{\alpha}^{T}\bm{\Psi}\bm{\nu}\\
&\text{s.t.}:\bm{s}+\mathscr{S}\bm{\Psi}\bm{\nu}\succeq  \bm{\Psi}\bm{\nu}\\
&\bm{0}\preceq \bm{s}, \quad (\bm{\mu}_{\bm{\alpha}}^{\bm{\pi}^{\ast}})^{T}\bm{s}\leq t,
\end{align}
where $t\geq 0$ is a constraint violation budget, and $\bm{\mu}_{\bm{\alpha}}^{\bm{\pi}^{\ast}}$ is the constraint violation distribution. Desai \textit{et al.} \cite{desai2012approximate} have shown that if $\bm{\nu}^{\ast}$ is a solution to the above problem, then
\begin{align}
\label{Eq:Tighter_Abbasi}
\expect_{s\sim \bm{\alpha}}[|v^{\ast}(s)-(\bm{\Psi}\bm{\nu}^{\ast})(s)|]\leq \inf_{\bm{\nu},\bm{u}\in \mathcal{U}} \dfrac{1}{u(s)}|v^{\ast}(s)-(\bm{\Psi}\bm{\nu})(s)|\left(\bm{\alpha}^{T}\bm{u}+\dfrac{2\bm{\mu}^{T}_{}\bm{u}(1+\beta_{\bm{u}})}{1-\gamma} \right),
\end{align}
where $\mathcal{U}\df \{\bm{u}\in \real^{n}: \bm{u}\preceq \bm{1}\}$. The upper bound in Eq. \eqref{Eq:Tighter_Abbasi} is tighter than that of Eq. \eqref{Eq:Tighter_Than_This} since $\mathcal{U}_{\bm{\Psi}}\subset \mathcal{U}$, and the right hand side of Eq. \eqref{Eq:Tighter_Abbasi} is smaller than that of Eq. \eqref{Eq:}. However, the results in \cite{desai2012approximate} requires idealized (stationary) distribution $\mu_{\alpha}^{\pi^{\ast}}$ for the s

Abbasi-Yadkori, \textit{et al}. \cite{abbasi2014linear} consider an approximate linear programming for the dual optimization problem considered in Eq. \eqref{Eq:dual_formulation}, and proposes a reduction from Markov decision problems to stochastic convex optimization. The main idea is to convert the dual ALP in Eq. \eqref{Eq:dual_formulation} into an unconstrained optimization problem via a penalty function method. In particular, consider the following function
\begin{align}
\label{Eq:Reduction}
f_{\omega}(\bm{\theta})\df \bm{c}^{T}\bm{\Psi}\bm{\theta}+\omega \|[\bm{\Psi}\bm{\theta}]_{-}\|_{1}+\omega \|\bm{\theta}^{T}\bm{\Psi}^{T}(\bm{P}-\bm{Q})\|_{1}.
\end{align}
where $\omega>0$ is a penalty factor term. Consider the iterations of the stochastic gradient descent $\bm{\theta}_{t+1}=\mathcal{P}_{\bm{\Theta}_{R}}(\bm{\theta}_{t}-\eta_{t}\bm{g}_{t})$, where $\bm{g}_{t}\in \partial f_{\omega}(\bm{\theta}_{t})$ is a sub-gradient vector of the function $f_{\omega}$, and $\bm{\Theta}_{R}^{d}\df \{\bm{\theta}\in \real^{d}:\|\bm{\theta}\|\leq R\}$ is the size of the policy class. It is shown that for any $\delta\in (0,1)$, with the probability of at least $1-\delta$, the following inequality holds
\begin{align}
\label{Eq:Performance_Bound}
\bm{c}^{T}\bm{\xi}^{\widehat{\bm{\theta}}_{T}}\leq \min_{\bm{\theta}\in \bm{\Theta}_{R}^{d}}\left(\bm{c}^{T}\bm{\xi}^{\bm{\theta}}+V_{1}(\bm{\theta})+V_{2}(\bm{\theta})+\mathcal{O}(\varepsilon)\right),
\end{align}
where $\widehat{\bm{\theta}}_{T}={1\over T}\sum_{t=1}^{T}\bm{\theta}_{t}$, the constraint violation functions are $V_{1}(\bm{\theta})\df \|[\bm{\Psi}\bm{\theta}]_{-}\|_{1}$, $V_{2}(\bm{\theta})\df \|\bm{\theta}^{T}\bm{\Psi}^{T}(\bm{P}-\bm{Q})\|_{1}$, and the  constants hidden in the asymptotic term are polynomials in $R$, $d$, $\log(1/\delta)$, $V_{1}(\bm{\theta})$,
$V_{2}(\bm{\theta})$, and $\tau_{\text{mix}}$. More recently, a similar reduction to the stochastic gradient descent has been considered by Banijamali, \textit{et al.} \cite{banijamali2018optimizing} for the mixture model studied in this paper. They have prove the following result
\begin{align}
\label{Eq:Performance_Bound_1}
\bm{c}^{T}\bm{\xi}^{\widehat{\bm{\theta}}_{T}}\leq \min_{\bm{\theta}\in \bm{\Theta}_{R}^{d}}\left(\bm{c}^{T}\bm{\xi}^{\bm{\theta}}+\left(\omega+\dfrac{6}{\omega(1-\gamma)}\right)V_{1}(\bm{\theta})+\dfrac{6\sqrt{m}CR }{1-\gamma}+\mathcal{O}\Bigg({1\over \omega}\Bigg)\right),
\end{align}
where once again, the constants hidden in the asymptotic term are polynomial in $R$, $m$, and $C$.

Notice that the performance bounds in Eqs \eqref{Eq:Performance_Bound} and \eqref{Eq:Performance_Bound_1} include the error terms due to the stationary and non-negativity constraints. In contrast, our performance bounds in Eq. \eqref{Eq:Bound_1} does not include such additional terms. Moreover, as we proved in Eq. \eqref{Eq:Constraint_Violation_Bound}, the error term due to the non-negativity constraint vanishes asymptotically in our proposed method.

\section{Proofs of Main Results}
\label{Section:Proof_of_Main_Results}
In this section, we prove the main results. We defer technical lemmas to appendices.

\subsection{Proof of Proposition \ref{Proposition:Convergence_Alg1}}
\label{Eq:Proof_Main_Proposition}

 First consider the steps to update the primal variables in Eq. \eqref{Eq:Recursion_1} of Algorithm \ref{Alg:1}. Recall from Eq. \eqref{Eq:Bregman_Projection_H0}-\eqref{Eq:Bregman_Projection_H} and the ensuing discussion in Section \ref{Subsection:MDP} that the equations \eqref{Eq:Recursion_1}-\eqref{Eq:Recursion_3} describing the update rules for the primal variables  in Algorithm \ref{Alg:1} are equivalent to the following two steps 
\begin{subequations}
	\begin{align}
	\label{Eq:first_equation}
	\nabla \phi(\widetilde{\bm{\theta}}_{t})&=\nabla \phi(\bm{\theta}_{t})-\eta_{t}\nabla_{\bm{\theta}}\widehat{\mathcal{L}}_{\bm{\Psi}}(\bm{\theta}_{t},\bm{\lambda}_{t}),\\
	\label{Eq:second_equation}
	\bm{\theta}_{t+1}&=\arg\min_{\bm{\theta}\in \bm{\Theta}_{2}}D_{\phi}(\bm{\theta};\widetilde{\bm{\theta}}_{t}),
	\end{align}
\end{subequations}
where we also recall the definition of the hyper-plane $\bm{\Theta}_{2}^{d}$ from Section \ref{Subsection:ALP}.

From Eq. \eqref{Eq:first_equation} and the characterization of the Bregman divergence in Definition \ref{Def:Bregman_divergence}, we obtain that
\small\begin{align}
\label{Eq:GLC1}
\eta_{t}\langle \nabla_{\bm{\theta}}\widehat{\mathcal{L}}_{\bm{\Psi}}(\bm{\theta}_{t},\bm{\lambda}_{t}),\bm{\theta}-\bm{\theta}_{t}\rangle=\langle \nabla \phi(\bm{\theta}_{t})-\nabla \phi(\widetilde{\bm{\theta}}_{t}),\bm{\theta}-\bm{\theta}_{t} \rangle.
\end{align}\normalsize
The next two lemmas generalize the law of cosines and Pythagorean inequality for Bregman divergences. The proofs of the following lemmas can be found in \cite{bubeck2015convex}:
\begin{lemma}\ \textsc{(Generalized Law of Cosine)}
	\label{Lemma:GLC}
	Let $\phi:\bm{\Theta}_{R}\rightarrow \real$ be a Legendre function. For all $\bm{\theta}\in \bm{\Theta}_{R}$ and $\bm{\theta}_{t},\widetilde{\bm{\theta}}_{t}\in \relint(\bm{\Theta}_{R})$, the following equality holds
	{\small	\begin{align}
		\label{Eq:GLC2}
		\langle \nabla \phi(\bm{\theta}_{t})-\nabla \phi(\widetilde{\bm{\theta}}_{t}),\bm{\theta}-\bm{\theta}_{t} \rangle
		=D_{\phi}(\bm{\theta};\bm{\theta}_{t})-D_{\phi}(\bm{\theta};\widetilde{\bm{\theta}}_{t})+D_{\phi}(\bm{\theta}_{t};\widetilde{\bm{\theta}}_{t}).
		\end{align}}
	\hfill $\square$
\end{lemma}

\begin{proposition}\textsc{(Generalized Pythagorean Inequality)}
	Let $\phi:\bm{\Theta}_{R}\rightarrow \real$ be a Legendre function. For all closed and convex sets $\bm{\Theta}_{2}\subseteq \real^{d}$ such that $\bm{\Theta}_{R}\cap \bm{\Theta}_{2}\not=\emptyset$, the following inequality holds
	\begin{align}
	\label{Eq:GPI}
	D_{\phi}\left(\Pi_{\bm{\Theta}_{2}}(\widetilde{\bm{\theta}}_{t});\widetilde{\bm{\theta}}_{t}\right)+D_{\phi}\left(\bm{\theta};\Pi_{\bm{\Theta}_{2}}(\widetilde{\bm{\theta}}_{t})\right)\leq D_{\phi}(\bm{\theta};\widetilde{\bm{\theta}}_{t}),
	\end{align}
	for all $\widetilde{\bm{\theta}}_{t}\in \bm{\Theta}_{R}$, where $	\Pi_{\bm{\Theta}_{2}}(\widetilde{\bm{\theta}}_{t})\df \arg\min_{\bm{\theta}\in \bm{\Theta}_{2}}D\left(\bm{\theta};\widetilde{\bm{\theta}}_{t}\right).$ \hfill $\square$
\end{proposition}

We proceed from Eq. \eqref{Eq:GLC1} by using Eq. \eqref{Eq:GLC2} of Lemma \ref{Lemma:GLC},
\small{\begin{align}
	\label{Eq:Inequality_01}
	&\eta_{t}\langle \nabla_{\bm{\theta}}\widehat{\mathcal{L}}_{\bm{\Psi}}(\bm{\theta}_{t},\bm{\lambda}_{t}),\bm{\theta}-\bm{\theta}_{t}\rangle=D_{\phi}(\bm{\theta};\bm{\theta}_{t})-D_{\phi}(\bm{\theta};\widetilde{\bm{\theta}}_{t})+D_{\phi}(\bm{\theta}_{t};\widetilde{\bm{\theta}}_{t}),
	\end{align}}\normalsize
Based on the generalized Pythagorean inequality \eqref{Eq:GPI} and the fact that $\bm{\theta}_{t+1}$ is the Bregman projection of $\widetilde{\bm{\theta}}_{t}$ onto the hyper-plane $\bm{\Theta}_{2}$ (see Eq. \eqref{Eq:second_equation}), we obtain 
\begin{align}
\label{Eq:Inequality_02}
D_{\phi}(\bm{\theta};\bm{\theta}_{t+1})\leq D_{\phi}(\bm{\theta};\widetilde{\bm{\theta}}_{t}).
\end{align}
Plugging Inequality \eqref{Eq:Inequality_02} into Eq. \eqref{Eq:Inequality_01} and rearranging the resulting terms results in
\small \begin{align}
\label{Eq:when_I_met_you}
D_{\phi}(\bm{\theta};\bm{\theta}_{t+1})\leq D_{\phi}(\bm{\theta};\bm{\theta}_{t})+D_{\phi}(\bm{\theta}_{t};\widetilde{\bm{\theta}}_{t})+\eta_{t}\langle \nabla_{\bm{\theta}}\widehat{\mathcal{L}}_{\bm{\Psi}}(\bm{\theta}_{t},\bm{\lambda}_{t}),\bm{\theta}_{t}-\bm{\theta}\rangle.
\end{align}\normalsize
To compute an upper bound on the divergence $D_{\phi}(\bm{\theta}_{t};\widetilde{\bm{\theta}}_{t})$, we leverage the definition of the Bregman divergence to obtain
\begin{align}
\nonumber
D_{\phi}(\bm{\theta}_{t};\widetilde{\bm{\theta}}_{t})+D_{\phi}(\widetilde{\bm{\theta}}_{t};\bm{\theta}_{t})
&=\langle \nabla \phi(\bm{\theta}_{t})-\nabla \phi(\widetilde{\bm{\theta}}_{t}),\bm{\theta}_{t}-\widetilde{\bm{\theta}}_{t} \rangle\\ \label{Eq:Upper_Bound}
&=\eta_{t}\langle \nabla_{\bm{\theta}} \widehat{\mathcal{L}}_{\bm{\Psi}}(\bm{\theta}_{t},\bm{\lambda}_{t}),\bm{\theta}_{t}-\widetilde{\bm{\theta}}_{t}\rangle,
\end{align}
where the last step follows by Eq. \eqref{Eq:first_equation}. Furthermore, using the Fenchel-Young inequality, we obtain
\small\begin{align}
\label{Eq:COMB_1}
\eta_{t}\langle \nabla_{\bm{\theta}} \widehat{\mathcal{L}}_{\bm{\Psi}}(\bm{\theta}_{t},\bm{\lambda}_{t}),\bm{\theta}_{t}-\widetilde{\bm{\theta}}_{t}\rangle\leq \dfrac{1}{2}\eta_{t}^{2}\|\nabla_{\bm{\theta}} \widehat{\mathcal{L}}_{\bm{\Psi}}(\bm{\theta}_{t},\bm{\lambda}_{t})\|_{\ast}^{2}+\dfrac{1}{2}\|\bm{\theta}_{t}-\widetilde{\bm{\theta}}_{t}\|^{2}.
\end{align}\normalsize
Now, since the modulus $\phi$ is a $1$-strongly convex modulus with respect to the induced norm $\|\cdot\|$ on $\bm{\Theta}_{2}$, the divergence $D_{\phi}(\cdot;\cdot)$ satisfies the following inequality
\begin{align}
\label{Eq:Lower_Bound_q}
D_{\phi}(\bm{\theta}_{t};\widetilde{\bm{\theta}}_{t})\geq \dfrac{1}{2}\|\bm{\theta}_{t}-\widetilde{\bm{\theta}}_{t}\|^{2}.
\end{align}
Combining Equations \eqref{Eq:Upper_Bound}, \eqref{Eq:COMB_1}, and \eqref{Eq:Lower_Bound_q} yields
\begin{align}
\label{Eq:upper_bound}
D_{\phi}(\bm{\theta}_{t};\widetilde{\bm{\theta}}_{t})\leq  \dfrac{1}{2}\eta_{t}^{2}\|\nabla_{\bm{\theta}} \widehat{\mathcal{L}}_{\bm{\Psi}}(\bm{\theta}_{t},\bm{\lambda}_{t})\|_{\ast}^{2}.
\end{align}
Substituting Inequality \eqref{Eq:upper_bound} in Eq. \eqref{Eq:when_I_met_you} and rearranging terms yields
\small \begin{align}
\nonumber
\eta_{t}\langle \nabla_{\bm{\theta}} \widehat{\mathcal{L}}_{\bm{\Psi}}(\bm{\theta}_{t},\bm{\lambda}_{t}),\bm{\theta}-\bm{\theta}_{t}\rangle &\leq D_{\phi}(\bm{\theta};\bm{\theta}_{t})-D_{\phi}(\bm{\theta};\bm{\theta}_{t+1})\\ \label{Eq:COMB_?}
&+\dfrac{1}{2}\eta_{t}^{2}\|\nabla_{\bm{\theta}} \widehat{\mathcal{L}}_{\bm{\Psi}}(\bm{\theta}_{t},\bm{\lambda}_{t})\|_{\ast}^{2}.
\end{align}\normalsize
We define the estimation error $\bm{e}_{t}\df \nabla_{\bm{\theta}} \widehat{\mathcal{L}}_{\bm{\Psi}}(\bm{\theta}_{t},\bm{\lambda}_{t})-\expect\left[\nabla_{\bm{\theta}}\widehat{\mathcal{L}}_{\bm{\Psi}}(\bm{\theta}_{t},\bm{\lambda}_{t})\big| \mathcal{F}_{t}\right]$, where $\mathcal{F}_{t}$ is the $\sigma$-field generated by the random variables $(\bm{\theta}_{k},\bm{\lambda}_{k})_{0\leq k\leq t-1}$. 
We sum both sides of Inequality \eqref{Eq:COMB_?} over $t=0,1,2,\cdots,T-1$,
\small{\begin{align}
	\nonumber
	&\sum_{t=0}^{T-1}\eta_{t}\langle \nabla_{\bm{\theta}} \mathcal{L}_{\bm{\Psi}}(\bm{\theta}_{t},\bm{\lambda}_{t}),\bm{\theta}-\bm{\theta}_{t}\rangle\leq  D_{\phi}(\bm{\theta};\bm{\theta}_{0})+\sum_{t=0}^{T-1}\eta_{t}\langle \bm{e}_{t},\bm{\theta}_{t}-\bm{\theta} \rangle
	\\ \label{Eq:COMBB_00}
	&+\sum_{t=0}^{T-1}\eta_{t}^{2}\|\nabla_{\bm{\theta}} \mathcal{L}_{\bm{\Psi}}(\bm{\theta}_{t},\bm{\lambda}_{t})\|_{\ast}^{2}+\sum_{t=0}^{T-1}\eta_{t}^{2}\|\bm{e}_{t}\|_{\ast}^{2},
	\end{align}}\normalsize
where in deriving Eq. \eqref{Eq:COMBB_00} we dropped $D_{\phi}(\bm{\theta};\bm{\theta}_{T})$ since it is non-negative. By the convexity of $\mathcal{L}_{\bm{\Psi}}(\cdot,\bm{\lambda}_{t})$, we have $\mathcal{L}_{\bm{\Psi}}(\bm{\theta}_{t},\bm{\lambda}_{t}),\bm{\theta}-\bm{\theta}_{t}\rangle \leq \mathcal{L}_{\bm{\Psi}}(\bm{\theta}_{t},\bm{\lambda}_{t})-\mathcal{L}_{\bm{\Psi}}(\bm{\theta},\bm{\lambda}_{t})$. Therefore,
\small {\begin{align}
	\nonumber
	&\sum_{t=0}^{T-1}\eta_{t}\mathcal{L}_{\bm{\Psi}}(\bm{\theta}_{t},\bm{\lambda}_{t})-\mathcal{L}_{\bm{\Psi}}(\bm{\theta},\bm{\lambda}_{t})\leq  D_{\phi}(\bm{\theta};\bm{\theta}_{0})+\sum_{t=0}^{T-1}\eta_{t}\langle \bm{e}_{t},\bm{\theta}_{t}-\bm{\theta} \rangle
	\\ \label{Eq:COMBB_1} &+\sum_{t=0}^{T-1}\eta_{t}^{2}\|\nabla_{\bm{\theta}} \mathcal{L}_{\bm{\Psi}}(\bm{\theta}_{t},\bm{\lambda}_{t})\|_{\ast}^{2}+\sum_{t=0}^{T-1}\eta_{t}^{2}\|\bm{e}_{t}\|_{\ast}^{2},
	\end{align}}\normalsize

From the recursion \eqref{Eq:Update_Rule_For_Lagrange_Multipliers} in Algorithm \ref{Alg:1}, for all $\bm{\lambda}\in \bm{\Lambda}_{\beta}^{nm}$ we have
\begin{align}
\nonumber
\|\bm{\lambda}_{t+1}-\bm{\lambda}\|_{2}&=\left\|\mathcal{P}_{\bm{\Lambda}_{\beta}^{nm}}(\bm{\lambda}_{t}+\eta_{t}\nabla_{\bm{\lambda}}\mathcal{L}_{\bm{\Psi}}(\bm{\theta}_{t},\bm{\lambda}_{t}) )-\mathcal{P}_{\bm{\Lambda}_{\beta}^{nm}}(\bm{\lambda})\right\|_{2}\\ \label{Eq:Square}
&\leq \ \left\|\bm{\lambda}_{t}+\eta_{t}\nabla_{\bm{\lambda}}\mathcal{L}_{\bm{\Psi}}(\bm{\theta}_{t},\bm{\lambda}_{t})-\bm{\lambda}\right\|_{2},
\end{align}
where the inequality is due to the non-expansive property of the projection, and $\mathcal{P}_{\bm{\Lambda}^{nm}_{\beta}}(\bm{x})=(G_{\beta}\cdot \bm{x})/(G_{\beta}\vee\|\bm{x}\|_{2})$ is the Euclidean projection onto the norm ball $\bm{\Lambda}$ defined after Lemma \ref{Lemma:1}. Squaring both sides of the inequality \eqref{Eq:Square} yields
\begin{align}
\nonumber
\|\bm{\lambda}_{t+1}-\bm{\lambda}\|_{2}^{2}=\|\bm{\lambda}_{t}-\bm{\lambda}\|_{2}^{2}+\eta_{t}^{2}\|\nabla_{\bm{\lambda}}\mathcal{L}_{\bm{\Psi}}(\bm{\theta}_{t},\bm{\lambda}_{t})\|_{2}^{2}-\eta_{t}\langle \nabla_{\bm{\lambda}}\mathcal{L}_{\bm{\Psi}}(\bm{\theta}_{t},\bm{\lambda}_{t}),\bm{\lambda}_{t}-\bm{\lambda} \rangle.
\end{align}
Taking the sum over $t=0,1,\cdots,T-1$ yields the following inequality
\small\begin{align}
\label{Eq:COMBB_2}
\sum_{t=0}^{T-1}\eta_{t}\langle \nabla_{\bm{\lambda}} \mathcal{L}_{\bm{\Psi}}(\bm{\theta}_{t},\bm{\lambda}_{t}),\bm{\lambda}_{t}-\bm{\lambda}\rangle\leq  {1\over 2}\|\bm{\lambda}\|_{2}^{2}+\sum_{t=0}^{T-1}\eta_{t}^{2}\|\nabla_{\bm{\lambda}} \mathcal{L}_{\bm{\Psi}}(\bm{\theta}_{t},\bm{\lambda}_{t})\|_{2}^{2}.
\end{align}\normalsize
where we used the fact that $\bm{\lambda}_{0}=\bm{0}$. 

Due to the convexity of $\mathcal{L}_{\bm{\Psi}}(\cdot,\bm{\lambda}_{t})$ and the concavity of $\mathcal{L}_{\bm{\Psi}}(\bm{\theta}_{t},\cdot)$, we have
\small\begin{subequations}
	\label{Eq:COMBB_3}
	\begin{align}
	\langle \nabla_{\bm{\theta}} \mathcal{L}_{\bm{\Psi}}(\bm{\theta}_{t},\bm{\lambda}_{t}),\bm{\theta}-\bm{\theta}_{t}\rangle &\leq \mathcal{L}_{\bm{\Psi}}(\bm{\theta}_{t},\bm{\lambda}_{t})-\mathcal{L}_{\bm{\Psi}}(\bm{\theta},\bm{\lambda}_{t})\\
	\langle  \nabla_{\bm{\lambda}}\mathcal{L}_{\bm{\Psi}}(\bm{\theta}_{t},\bm{\lambda}_{t}),\bm{\lambda}_{t}-\bm{\lambda} \rangle &\leq  \mathcal{L}_{\bm{\Psi}}(\bm{\theta}_{t},\bm{\lambda})-\mathcal{L}_{\bm{\Psi}}(\bm{\theta}_{t},\bm{\lambda}_{t}).
	\end{align}
\end{subequations}\normalsize
Combining Eqs. \eqref{Eq:COMBB_1}, \eqref{Eq:COMBB_2}, and \eqref{Eq:COMBB_3} thus results in the following inequality
\footnotesize\begin{align}
\label{Eq:UPPER}
\sum_{t=0}^{T-1}\eta_{t}(\mathcal{L}_{\bm{\Psi}}(\bm{\theta}_{t},\bm{\lambda})-\mathcal{L}_{\bm{\Psi}}(\bm{\theta},\bm{\lambda}_{t}))&\leq D_{\phi}(\bm{\theta};\bm{\theta}_{0})+{1\over 2}\|\bm{\lambda}\|_{2}^{2}+\sum_{t=0}^{T-1}\eta_{t}\langle \bm{e}_{t},\bm{\theta}_{t}-\bm{\theta} \rangle+\sum_{t=0}^{T-1}\eta_{t}^{2}\|\bm{e}_{t}\|_{\ast}^{2}\\ \nonumber
&+\sum_{t=0}^{T-1}\eta_{t}^{2}\|\nabla_{\bm{\theta}} \mathcal{L}_{\bm{\Psi}}(\bm{\theta}_{t},\bm{\lambda}_{t})\|_{\ast}^{2}+\sum_{t=0}^{T-1}\eta_{t}^{2}\|\nabla_{\bm{\lambda}} \mathcal{L}_{\bm{\Psi}}(\bm{\theta}_{t},\bm{\lambda}_{t})\|_{2}^{2}.
\end{align}\normalsize
In the next lemma, we prove the following concentration inequalities for the terms involving the estimation errors:
\begin{lemma} \textsc{(Concentration Inequalities for Estimation Error)}
	\label{Lemma:1}
	For all $\bm{\theta}_{t},\bm{\theta}\in \bm{\Theta}_{R}^{d}$ and $t\in [T]$, the random variable $\langle \bm{e}_{t},\bm{\theta}_{t}-\bm{\theta}\rangle$ is sub-Gaussian with the Orlicz norm of  $\|\langle\bm{e}_{t},\bm{\theta}_{t}-\bm{\theta}\rangle \|_{\psi_{2}}\leq 8R^{2}C_{e}^{2}$. Moreover,
	\small	\begin{subequations}
		\label{Eq:Concentration_1}
		\begin{align}
		\label{Eq:CON_11}
		\prob\left[\left|\sum_{t=0}^{T-1}\eta_{t}\langle \bm{e}_{t},\bm{\theta}_{t}-\bm{\theta}\rangle\right|\geq \delta\Bigg| \mathcal{F}_{t} \right]\leq 2\exp\left({-c_{0}\cdot\dfrac{\delta^{2}}{8R^{2}C_{e}^{2}\sum_{t=0}^{T-1}\eta_{t}^{2}}}\right),
		\end{align}\normalsize
	\end{subequations}
	where $C_{e}>0$ is the constant defined in Eq. \eqref{Eq:Constant_C}, and $c_{0}>0$ is a universal constant.

In addition, the random variable $\|\bm{e}_{t}\|_{\ast}^{2}$ is a sub-exponential with the Orlicz norm of $\| \|\bm{e}_{t}\|_{\ast}^{2} \|_{\psi_{1}}\leq \sqrt{{\log(2)\over 2}}C^{2}(n)C_{e}^{2}\leq C^{2}(n)C_{e}^{2}$ satisfying the following concentration inequality
	\begin{subequations}
		\label{Eq:Concentration_2}
	 \begin{align}
		\label{Eq:CONNN1}
	&\prob\Bigg[\sum_{t=0}^{T-1}\eta_{t}^{2}(\|\bm{e}_{t}\|_{2}^{2}-\expect[\|\bm{e}_{t}\|_{2}^{2}|\mathcal{F}_{t}])\geq \delta \Bigg| \mathcal{F}_{t}   \Bigg] \\ \nonumber &\leq 2\exp\left(-c_{1}\cdot \min\left\{ \dfrac{\delta^{2}}{C_{e}^{2}C^{2}(n)\sum_{t=0}^{T-1}\eta_{t}^{4}},\dfrac{\delta}{C_{e}C(n)\sum_{t=0}^{T-1}\eta_{t}^{2}} \right\} \right),
		\end{align}
		$c_{1}>0$ is a universal constant and $C(n)$ is a constant depending on $n$ only such that $\|\bm{v}\|\leq C(n)\|\bm{v}\|_{2}$ for all $\bm{v}\in \real^{n}$.\hfill $\square$
	\end{subequations}
\end{lemma}
We use the concentration bounds in Eqs. \eqref{Eq:Concentration_1} and \eqref{Eq:Concentration_2} in conjunction with the upper bound in Eq. \eqref{Eq:UPPER}. Taking the union bound, we obtain that with the probability of (at least) $1-\rho$ the following inequality holds
\small \begin{align}
\label{Eq:R.H.S}
&\sum_{t=0}^{T-1}\eta_{t}(\mathcal{L}_{\bm{\Psi}}(\bm{\theta}_{t},\bm{\lambda})-\mathcal{L}_{\bm{\Psi}}(\bm{\theta},\bm{\lambda}_{t}))\leq D_{\phi}(\bm{\theta};\bm{\theta}_{0})+{1\over 2}\|\bm{\lambda}\|_{2}^{2}
\\  \nonumber
&+\sum_{t=0}^{T-1}\eta_{t}^{2}(\|\nabla_{\bm{\theta}} \mathcal{L}_{\bm{\Psi}}(\bm{\theta}_{t},\bm{\lambda}_{t})\|_{\ast}^{2}+\|\nabla_{\bm{\lambda}} \mathcal{L}_{\bm{\Psi}}(\bm{\theta}_{t},\bm{\lambda}_{t})\|_{2}^{2}+\expect[\|\bm{e}_{t}\|_{\ast}^{2}|\mathcal{F}_{t}])\\ \nonumber
&+2\sqrt{2}{RC_{e}\over \sqrt{c}_{0}}\left(\ln\left({4\over \rho}\right)\sum_{t=0}^{T-1}\eta_{t}^{2} \right)^{1\over 2}+{C_{e}C(n)}\max\left\{\left({1\over c_{1}}\ln\left({4\over \rho}\right)\sum_{t=0}^{T-1}\eta_{t}^{4} \right)^{1\over 2},{1\over c_{1}}\ln\left({4\over \rho}\right)\sum_{t=0}^{T-1}\eta_{t}^{2}\right\}.
\end{align}\normalsize
Divide both sides of the inequality by ${1\over \sum_{t=0}^{T-1}\eta_{t}}$ and recall the definitions of the running averages $\widehat{\bm{\theta}}_{T}$ and $\widehat{\bm{\lambda}}_{T}$ from Eq. \eqref{Eq:Running_averages} to obtain
\begin{align}
\label{Eq:Lagrangian_Left_Hand_Side}
\hspace{-3mm}\mathcal{L}_{\bm{\Psi}}(\widehat{\bm{\theta}}_{T},\bm{\lambda})-\mathcal{L}_{\bm{\Psi}}(\bm{\theta},\widehat{\bm{\lambda}}_{T})\leq \dfrac{1}{\sum_{t=0}^{T-1}\eta_{t}}\left[\text{r.h.s. of Eq. \ } \eqref{Eq:R.H.S} \right].
\end{align}\normalsize
We expand the Lagrangian functions on the left hand side of Eq. \eqref{Eq:Lagrangian_Left_Hand_Side}, using the definition of the Lagrangian function in Eq. \eqref{Eq:Lagrangian}. Furthermore, since the inequality \eqref{Eq:Lagrangian_Left_Hand_Side} holds for all $\bm{\theta}\in \bm{\Theta}_{2}^{d}\cap \bm{\Theta}_{R}^{d}$, we set $\bm{\theta}=\bm{\theta}^{\ast}$. We obtain that
\small \begin{align}  
\nonumber 
&\bm{c}^{T}\bm{\Psi}\widehat{\bm{\theta}}_{T}-\left[\bm{\lambda}^{T}\bm{\Psi}\widehat{\bm{\theta}}_{T}+\dfrac{1}{2\sum_{t=0}^{T-1}\eta_{t}}\|\bm{\lambda}\|_{2}^{2}\right]
-\bm{c}^{T}\bm{\Psi}\bm{\theta}^{\ast}+\widehat{\bm{\lambda}}^{T}_{T}\bm{\Psi}\bm{\theta}^{\ast}\\
& \label{Eq:Leading_To_Equation}
\leq  \dfrac{1}{\sum_{t=0}^{T-1}\eta_{t}}  \Bigg[ D_{\phi}(\bm{\theta}^{\ast};\bm{\theta}_{0})+\sum_{t=0}^{T-1}\eta_{t}^{2}\expect[\|\bm{e}_{t}\|_{\ast}^{2}|\mathcal{F}_{t}]+\sum_{t=0}^{T-1}\eta_{t}^{2}(\|\nabla_{\bm{\theta}} \mathcal{L}_{\bm{\Psi}}(\bm{\theta}_{t},\bm{\lambda}_{t})\|_{\ast}^{2}+\|\nabla_{\bm{\lambda}} \mathcal{L}_{\bm{\Psi}}(\bm{\theta}_{t},\bm{\lambda}_{t})\|_{2}^{2})\\ \label{Eq:In_Front_Of}
&+2\sqrt{2}{RC_{e}\over c_{0}}\left(\ln\left({4\over \rho}\right)\sum_{t=0}^{T-1}\eta_{t}^{2} \right)^{1\over 2}+{C_{e}C(n)}\max\left\{\left({1\over c_{1}}\ln\left({4\over \rho}\right)\sum_{t=0}^{T-1}\eta_{t}^{4} \right)^{1\over 2},{1\over c_{1}}\ln\left({4\over \rho}\right)\sum_{t=0}^{T-1}\eta_{t}^{2}\right\}\Bigg].
\end{align}\normalsize
Since the preceding inequality holds for all Lagrange multipliers $\bm{\lambda}\in \bm{\Lambda}_{\beta}^{nm}\df \{\bm{\lambda}\in \real_{+}^{nm}:\|\bm{\lambda}\|_{2}\leq G_{\beta} \}$, we can maximize the terms inside the bracket on the left hand side by letting 
\begin{align}
\bm{\lambda}= -\dfrac{G_{\beta}\cdot \left({1\over 2}\left(\sum_{t=0}^{T-1}\eta_{t}\right)[\bm{\Psi}\widehat{\bm{\theta}}_{T}]_{-}\right)}{G_{\beta}\vee  \left\|{1\over 2}\left(\sum_{t=0}^{T-1}\eta_{t}\right)[\bm{\Psi}\widehat{\bm{\theta}}_{T}]_{-}\right\|_{2}}. 
\end{align}
From Eq. \eqref{Eq:In_Front_Of}, we then derive that
\small\begin{align}
\nonumber
&\bm{c}^{T}\bm{\Psi}\widehat{\bm{\theta}}_{T}-\bm{c}^{T}\bm{\Psi}\bm{\theta}^{\ast}+\widehat{\bm{\lambda}}^{T}_{T}\bm{\Psi}\bm{\theta}^{\ast}+\left[\dfrac{2G_{\beta} V^{2}(\widehat{\bm{\theta}}_{T})\left(\sum_{t=0}^{T-1}\eta_{t}\right)}{G_{\beta}\vee  \left(\sum_{t=0}^{T-1}\eta_{t}\right)V(\widehat{\bm{\theta}}_{T}) }-\dfrac{G_{\beta}^{2} V^{2}(\widehat{\bm{\theta}}_{T})\left(\sum_{t=0}^{T-1}\eta_{t}\right)}{G_{\beta}^{2}\vee  \left(\sum_{t=0}^{T-1}\eta_{t}\right)^{2}V^{2}(\widehat{\bm{\theta}}_{T})} \right] \\ 
\label{Eq:Left_Hand_Side} &\leq \text{r.h.s. of Eq. \ }\eqref{Eq:In_Front_Of},
\end{align}\normalsize
where to write the inequality, we also used the definition of the constraint violation $V(\widehat{\bm{\theta}}_{T})\df {1\over 2} \big\|\big[\bm{\Psi}\widehat{\bm{\theta}}_{T}\big]_{-}\big\|_{2} $.

We now notice that $\widehat{\bm{\lambda}}_{T}^{T}\bm{\Psi}\bm{\theta}^{\ast}\geq 0$ since $\bm{0}\preceq \widehat{\bm{\lambda}}_{T}$ and $\bm{0}\preceq \bm{\Psi}\bm{\theta}^{\ast}$ as the optimal solution $\bm{\theta}^{\ast}$ must be feasible.  In addition, 
\begin{align}
\dfrac{2G_{\beta} V^{2}(\widehat{\bm{\theta}}_{T})\left(\sum_{t=0}^{T-1}\eta_{t}\right)}{G_{\beta}\vee  \left(\sum_{t=0}^{T-1}\eta_{t}\right)V(\widehat{\bm{\theta}}_{T}) }-\dfrac{G_{\beta}^{2} V^{2}(\widehat{\bm{\theta}}_{T})\left(\sum_{t=0}^{T-1}\eta_{t}\right)}{G_{\beta}^{2}\vee  \left(\sum_{t=0}^{T-1}\eta_{t}\right)^{2}V^{2}(\widehat{\bm{\theta}}_{T})}\geq 0.
\end{align}
We drop the non-negative term from the left hand  of Eq. \eqref{Eq:Left_Hand_Side}. We derive
\begin{align}
\label{Eq:Left_Hand_Side_1}
&\bm{c}^{T}\bm{\Psi}\widehat{\bm{\theta}}_{T}-\bm{c}^{T}\bm{\Psi}\bm{\theta}^{\ast}\leq \text{r.h.s. of Eq. \ }\eqref{Eq:In_Front_Of}.
\end{align}
Now, we upper the deterministic gradients as follows
\begin{align}
\label{Eq:Inequalities_11}
\left\|\nabla_{\bm{\lambda}} \mathcal{L}_{\bm{\Psi}}(\bm{\theta}_{t},\bm{\lambda}_{t})\right\|_{2}\leq  \|\bm{\Psi}\bm{\theta}_{t}\|_{2}\leq \sigma_{\max}(\bm{\Psi})\|\bm{\theta}_{t}\|_{2}\stackrel{\rm{(a)}}{\leq} 2R\sigma_{\max}(\bm{\Psi}),
\end{align}
where the last inequality is due to  the fact that $\text{dom}(\phi)= \bm{\Theta}_{R}$ and thus $\bm{\theta}_{t}\in \bm{\Theta}_{R}$. Similarly, we have
\begin{align}
\big\|\nabla_{\bm{\theta}}\mathcal{L}_{\bm{\Psi}}(\bm{\theta}_{t},\bm{\lambda}_{t})\big\|_{\ast}=\big\|\bm{c}^{T}\bm{\Psi}+\bm{\lambda}_{t}^{T}\bm{\Psi}\big\|_{\ast}\leq \|\bm{c}^{T}\bm{\Psi}\|_{\ast}+\|\bm{\lambda}_{t}^{T}\bm{\Psi}\|_{\ast} \label{Eq:Inequalities_112}
\leq (c_{\max}\sqrt{nm}+G_{\beta})\|\bm{\Psi}\|_{\ast, 2}.
\end{align} 
where in the last inequality, $\|\cdot\|_{\ast,2}$  is the subordinate norm of the matrix $\bm{\Psi}$, and we used the fact that $\|\bm{\lambda}_{t}\|_{2}\leq G_{\beta}$, since $\bm{\lambda}_{t}\in \bm{\Lambda}_{\beta}^{nm}$.$\hfill \blacksquare$

We substitute the upper bounds \eqref{Eq:Inequalities_11} and \eqref{Eq:Inequalities_112} into Eq. \eqref{Eq:Left_Hand_Side_1},
\small{\begin{align}
\label{Eq:Piano}
&\bm{c}^{T}\bm{\Psi}\widehat{\bm{\theta}}_{T}-\bm{c}^{T}\bm{\Psi}\bm{\theta}^{\ast} \nonumber
\leq \dfrac{1}{\sum_{t=0}^{T-1}\eta_{t}} \Bigg[D_{\phi}(\bm{\theta}^{\ast};\bm{\theta}_{0})+\left(2R\sigma_{\max}(\bm{\Psi})+(c_{\max}\sqrt{nm}+S)\|\bm{\Psi}\|_{\ast,2}+C_{e}^{2} \right)\left(\sum_{t=0}^{T-1}\eta_{t}^{2}\right)\\ \nonumber 
&+2\sqrt{2}{RC_{e}\over c_{0}}\left(\ln\left({4\over \rho}\right)\sum_{t=0}^{T-1}\eta_{t}^{2} \right)^{1\over 2}+\dfrac{C_{e}C(n)}{c_{1}}\max\Bigg\{\Bigg(\ln\Bigg({4\over \rho}\Bigg)\sum_{t=0}^{T-1}\eta_{t}^{4} \Bigg)^{1\over 2},\ln\Bigg({4\over \rho}\Bigg)\sum_{t=0}^{T-1}\eta_{t}^{2}\Bigg)\Bigg\} \Bigg],
\end{align}}\normalsize 
where we used the fact that $\expect[\|\bm{e}_{t}\|_{\ast}^{2}|\mathcal{F}_{t}]\leq C_{e}^{2}$ as $\|\bm{e}_{t}\|_{\ast}\leq C_{e}$ due to Eq. \eqref{Eq:Ce}.

We choose the diminishing learning as $\eta_{t}={1\over \sqrt{t+1}}$. Using the Riemann sum approximation in conjunction with this learning rate yields the following inequalities
\begin{subequations}
	\begin{align}
	\sum_{t=0}^{T-1}\eta_{t}&\geq \int_{0}^{T-1}\hspace{-3mm}\dfrac{\mathrm{d}t}{\sqrt{t+1}} \geq 2(\sqrt{T}-1),\quad	\sum_{t=0}^{T-1}\eta_{t}\leq 1+\int_{0}^{T-1}\hspace{-3mm} \dfrac{\mathrm{d}t }{\sqrt{t+1}}\leq 2\sqrt{T}-1,    \\ 
	\sum_{t=0}^{T-1}\eta_{t}^{2}&\leq 1+ \int_{0}^{T-1}\hspace{-3mm}\dfrac{\mathrm{d}t}{t+1} \leq 4\log(T),  \quad
	\sum_{t=0}^{T-1}\eta_{t}^{4}\leq  1+\int_{0}^{T-1}\hspace{-4mm}\dfrac{\mathrm{d}t}{(t+1)^{2}} \leq 2-\dfrac{1}{T}\leq 2.
	\end{align}
\end{subequations}
Plugging this upper and lower bounds in Eq. \eqref{Eq:Piano} completes the proof of the high probability bound in \eqref{Eq:Bound_1} of Proposition \ref{Proposition:Convergence_Alg1}.

To proof of the in expectation bound in Eq. \eqref{Eq:Bound_2} of Proposition \ref{Proposition:Convergence_Alg1}, we begin from Eq. \eqref{Eq:UPPER}. By taking the conditional expectation $\expect[\cdot |\mathcal{F}_{t}]$ from both sides of the inequality and recall that $\expect[\langle \bm{e}_{t},\bm{\theta}_{t}-\bm{\theta}^{\ast}\rangle|\mathcal{F}_{t}]=0$. Therefore, we obtain
\begin{align}
\nonumber
\sum_{t=0}^{T-1}\eta_{t}\expect[(\mathcal{L}_{\bm{\Psi}}(\bm{\theta}_{t},\bm{\lambda})&-\mathcal{L}_{\bm{\Psi}}(\bm{\theta}^{\ast},\bm{\lambda}_{t}))|\mathcal{F}_{t}]\leq D_{\phi}(\bm{\theta}^{\ast};\bm{\theta}_{0})+{1\over 2}\|\bm{\lambda}\|_{2}^{2}+\sum_{t=0}^{T-1}\eta_{t}^{2}\expect[\|\bm{e}_{t}\|_{\ast}^{2}|\mathcal{F}_{t}]\\ \nonumber
&+\sum_{t=0}^{T-1}\eta_{t}^{2}\expect[\|\nabla_{\bm{\theta}} \mathcal{L}_{\bm{\Psi}}(\bm{\theta}_{t},\bm{\lambda}_{t})\|_{\ast}^{2}|\mathcal{F}_{t}]+\sum_{t=0}^{T-1}\eta_{t}^{2}\expect[\|\nabla_{\bm{\lambda}} \mathcal{L}_{\bm{\Psi}}(\bm{\theta}_{t},\bm{\lambda}_{t})\|_{2}^{2}|\mathcal{F}_{t}].
\end{align}
Moving ${1\over 2}\|\bm{\lambda}\|_{2}^{2}$ from the right hand side to the left hand side and using the expansion in Eq. \eqref{Eq:In_Front_Of} yields
\begin{align}
\label{Eq:In_expectation_bound}
&\sum_{t=0}^{T-1}\eta_{t}\left(\expect[\bm{c}^{T}\bm{\Psi}\widehat{\bm{\theta}}_{T}|\mathcal{F}_{t}]-\bm{c}^{T}\bm{\Psi}\bm{\theta}^{\ast}\right)\leq D_{\phi}(\bm{\theta}^{\ast};\bm{\theta}_{0})+\sum_{t=0}^{T-1}\eta_{t}^{2}\expect[\|\bm{e}_{t}\|_{\ast}^{2}|\mathcal{F}_{t}]\\ \nonumber
&+\sum_{t=0}^{T-1}\eta_{t}^{2}\expect[\|\nabla_{\bm{\theta}} \mathcal{L}_{\bm{\Psi}}(\bm{\theta}_{t},\bm{\lambda}_{t})\|_{\ast}^{2}|\mathcal{F}_{t}]+\sum_{t=0}^{T-1}\eta_{t}^{2}\expect[\|\nabla_{\bm{\lambda}} \mathcal{L}_{\bm{\Psi}}(\bm{\theta}_{t},\bm{\lambda}_{t})\|_{2}^{2}|\mathcal{F}_{t}].
\end{align}
Plugging the upper bounds for each terms on the right hand side and taking another expectation with respect to the filtration $\mathcal{F}_{t}$ completes the proof of Inequality \eqref{Eq:Bound_2}.

Now, we prove the asymptotic bound \eqref{Eq:Constraint_Violation_Bound} on the constraint violation. To do so, consider two scenarios. Namely, ${G_{\beta}\over \sum_{t=0}^{T-1}\eta_{t}}> V(\widehat{\bm{\theta}}_{T})$, and ${G_{\beta}\over \sum_{t=0}^{T-1}\eta_{t}}\leq V(\widehat{\bm{\theta}}_{T})$.
In the first case, the constraint violation has the following upper bound
\begin{align}
\label{Eq:Obtain_1}
V(\widehat{\bm{\theta}}_{T})< \dfrac{G_{\beta}}{2(\sqrt{T}-1)}=\dfrac{1}{2(1-\gamma)(\sqrt{T}-1)}\vee \dfrac{\beta}{2(\sqrt{T}-1) }.
\end{align}
In the second case, from Eq. \eqref{Eq:Left_Hand_Side}, we obtain that 
\begin{align}
\label{Eq:Lower_Bound}
&\bm{c}^{T}\bm{\Psi}\widehat{\bm{\theta}}_{T}-\bm{c}^{T}\bm{\Psi}\bm{\theta}^{\ast}+\widehat{\bm{\lambda}}^{T}_{T}\bm{\Psi}\bm{\theta}^{\ast}+2G_{\beta}V(\widehat{\bm{\theta}}_{T})-\dfrac{G_{\beta}^{2}}{\sum_{t=0}^{T-1}\eta_{t}} \leq \alpha_{T},
\end{align} 
where $\alpha_{T}$ encapsulates all the terms on the right hand side, and $\alpha_{T}=\mathcal{O}(1/\sqrt{T})$.

We compute a lower bound on the first two terms of Eq. \eqref{Eq:Lower_Bound} as follows
\begin{align}
\nonumber
\bm{c}^{T}\bm{\Psi}\widehat{\bm{\theta}}_{T}-\bm{c}^{T}\bm{\Psi}\bm{\theta}^{\ast}&\stackrel{\rm{(a)}}{\geq} -\|\bm{c}\bm{\Psi}\|_{\ast}\|\widehat{\bm{\theta}}_{T}-\bm{\theta}^{\ast}\|\\ \nonumber
&\stackrel{\rm{(b)}}{\geq} -2R\|\bm{c}\bm{\Psi}\|_{\ast}\\ \nonumber
&\stackrel{\rm{(c)}}{\geq} -2R\|\bm{c}\|_{2}\|\bm{\Psi}\|_{\ast,2}\\ \label{Eq:LL1}
&\stackrel{\rm{(d)}}{\geq} -2Rc_{\max}\sqrt{nm}\|\bm{\Psi}\|_{\ast,2},
\end{align}
where $\rm{(a)}$ and $\rm{(c)}$ follows from the Cauchy-Schwarz inequality,  $\rm{(b)}$ from the fact that $\widehat{\bm{\theta}}_{T},\bm{\theta}^{\ast}\in \bm{\Theta}_{R}^{d}$, and $\rm{(d)}$ from the fact that $c_{a}(s)\in [c_{\min},c_{\max}]$ for all $(a,s)\in \mathcal{A}\times \mathcal{S}$. The third term on the left hand side of Eq. \eqref{Eq:Lower_Bound} can be dropped since it is non-negative.

From Eq. \eqref{Eq:LL1}, we thus obtain that
\small{\begin{align}
	\label{Eq:Obtain_2}
	\hspace{-3mm}V(\widehat{\bm{\theta}}_{T})\leq  &\dfrac{1}{G_{\beta}}\bigg(Rc_{\max}\sqrt{nm}\|\bm{\Psi}\|_{\ast,2}+{1\over 2}\alpha_{T}\bigg)+\dfrac{G_{\beta}}{4(\sqrt{T}-1)}. 
	\end{align}}\normalsize
For a sufficiently large hyper-parameter $\beta>(1-\gamma)^{-1}$, $G_{\beta}=\beta$ and the upper bounds in Eqs. \eqref{Eq:Obtain_1}-\eqref{Eq:Obtain_2} can be aggregated into a unified bound as below
\small{\begin{align}
	V(\widehat{\bm{\theta}}_{T})\leq {Rc_{\max}\sqrt{nm}\|\bm{\Psi}\|_{\ast,2}\over \beta}+{1\over 2\beta}\alpha_{T}+\dfrac{\beta}{2(\sqrt{T}-1) }.\qedb
	\end{align}}\normalsize

\section{Numerical Experiments}
\label{Section:Numerical Experiments}
In this section, we discuss numerical experiments involving the application of Algorithm \ref{Alg:1} to a queueing problem. Following the work of de Farias and Van Roy  \cite{de2003linear}, in the following queuing model we assume that at most one event (arrival/departure) occurs at each time step. 

\subsection{Single Queue with Controlled Service Rate}

\begin{figure}[t!]
	\vspace{10mm}
	\begin{center}
		\subfigure{
			\includegraphics[trim={3cm 3cm 3cm 3cm}, width=.28\linewidth]{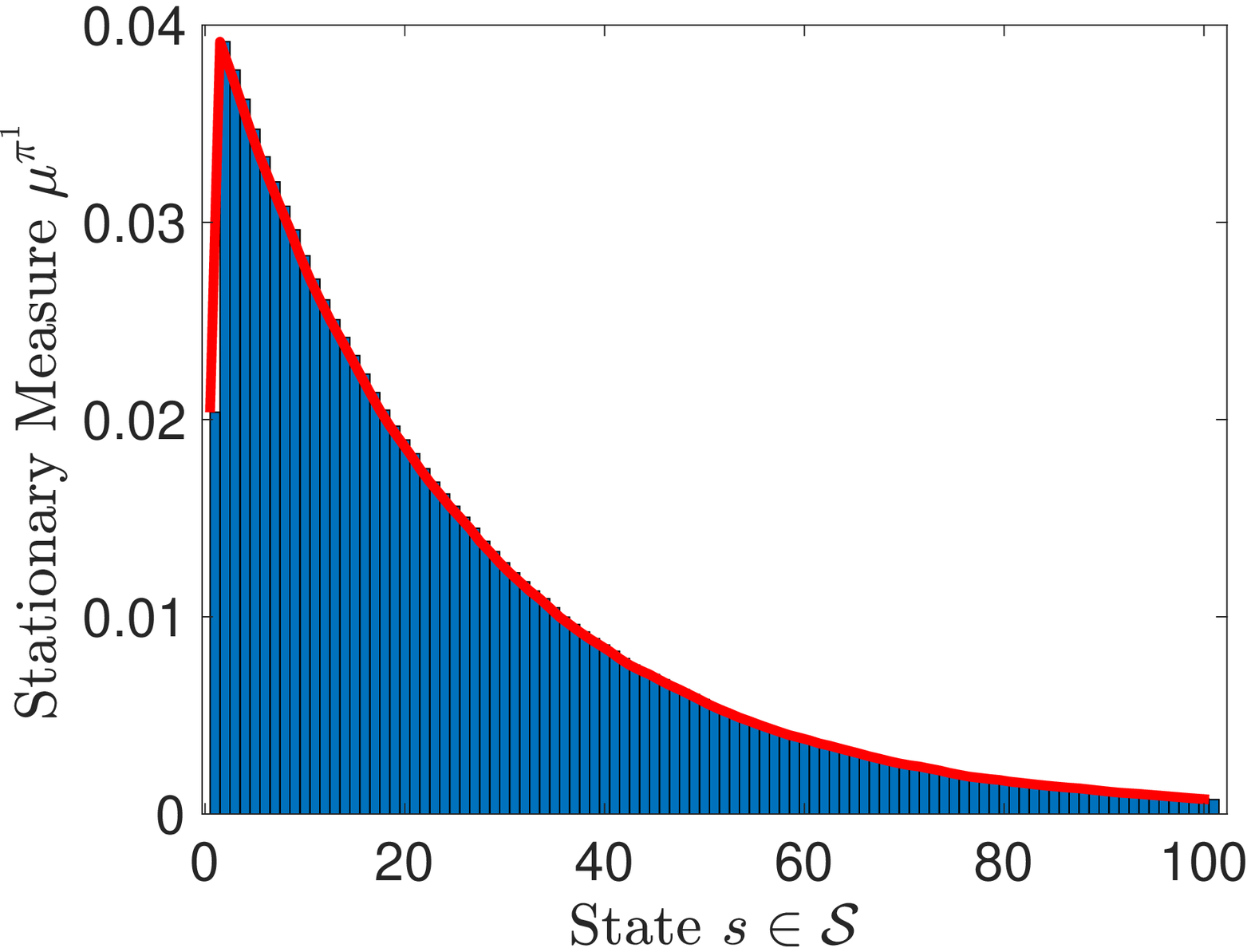}\hspace{20mm}
			\includegraphics[trim={3cm 3cm 3cm 3cm}, width=.28\linewidth]{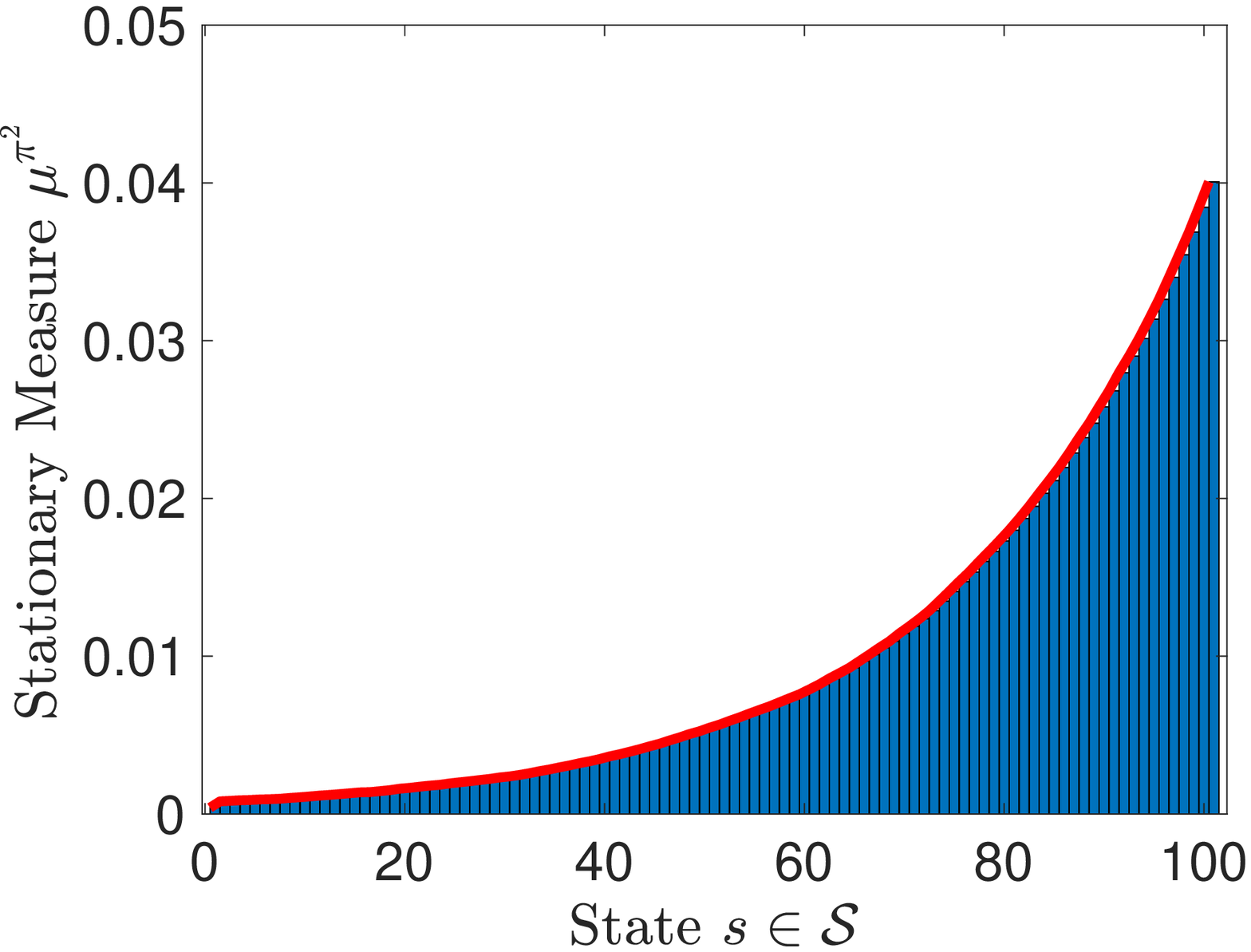}
		}
	\vspace{4mm}
	\end{center}
\vspace{4mm}
	\caption{\footnotesize{The histogram of stationary measures $\bm{\mu}^{\bm{\pi}_{1}}$ and $\bm{\mu}^{\bm{\pi}_{2}}$ associated with the base policies $\bm{\pi}^{1}$ (left) and $\bm{\pi}^{2}$ (right), evaluated empirically by computing the frequency of visits to each state during $T=5\times 10^{7}$ epochs.} }
	\label{Fig:1}
\end{figure}

The queuing model we describe here has been studied by de Farias and Van Roy \cite{de2003linear} and more recently by Banijamali, \textit{et al}. \cite{banijamali2018optimizing}, and consists of a single queue with the buffer size of $L=100$ without the discount factor, \textit{i.e.}, $\gamma=1$. We assume that the jobs arrive at the queue with the probability $\varrho=0.5$ in any unit of time. Service rates or departure probabilities $a_{t}$ are chosen from the set of actions $\mathcal{A}=\{0.2,0.4,0.6,0.8\}$. Therefore, the length (state) of the queue at each time $t\in [T]$ is determined by
\begin{align}
s_{t+1}=\left\{
\begin{array}{ll}
s_{t}+1 & \mbox{with probability}\ \varrho \\
s_{t}-1 & \mbox{with probability}\ a_{t}\\
s_{t} & \mbox{otherwise}
\end{array}
\right.
\end{align}
Here, $s_{t}\in \mathcal{S}\df \{1,2,\cdots,L\}$ for all $t\in [T]$. Associated with the action $a_{t}\in \mathcal{A}$ and the state $s_{t}\in \mathcal{S}$. Given the samples, $(s_{t},a_{t})$, the learner observes the cost $c(s_{t},a_{t})$ according to the following equation
\begin{align}
c(s_{t},a_{t})=s_{t}^{2}+60a_{t}^{3}.
\end{align}
Following the work of \cite{banijamali2018optimizing}, we consider two base policies independent of the states. More formally, we consider  
\begin{subequations}
	\label{Eq:Policies_12}
	\begin{align}
	\bm{\pi}^{1}(s)&=[0.25,.25,.25,.25],\\
	\bm{\pi}^{2}(s)&=[0.3,0.3,0.2,0.2], \quad \forall s\in \mathcal{S}.  
	\end{align}
\end{subequations}
In Figure \ref{Fig:1}, we illustrate the stationary measures $\bm{\mu}^{\bm{\pi}^{i}}$ associated with the base policies $\bm{\pi}^{i}$ for $i=1,2,$. To empirically compute these measures, we enumerated the number of visits to each state, and computed the frequencies of visits of each state during $T=5\times 10^{7}$ epochs. 

\begin{figure}[t!]
	\begin{center}x
		\subfigure{
			\includegraphics[trim={3cm 3cm 3cm 3cm}, width=.3\linewidth]{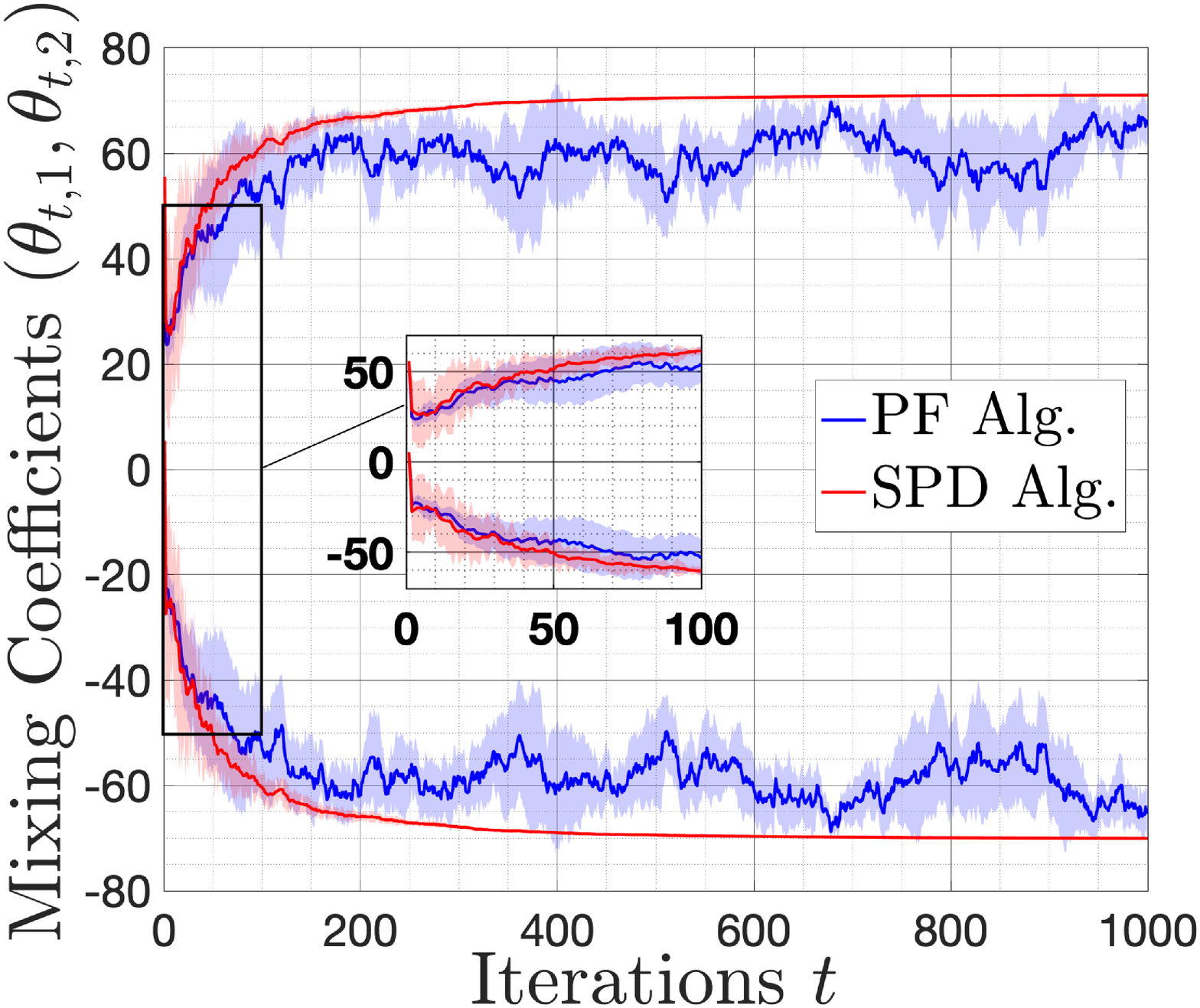}} \hspace{14mm}
		\subfigure{
			\includegraphics[trim={3cm 3cm 3cm 3cm}, width=.3\linewidth]{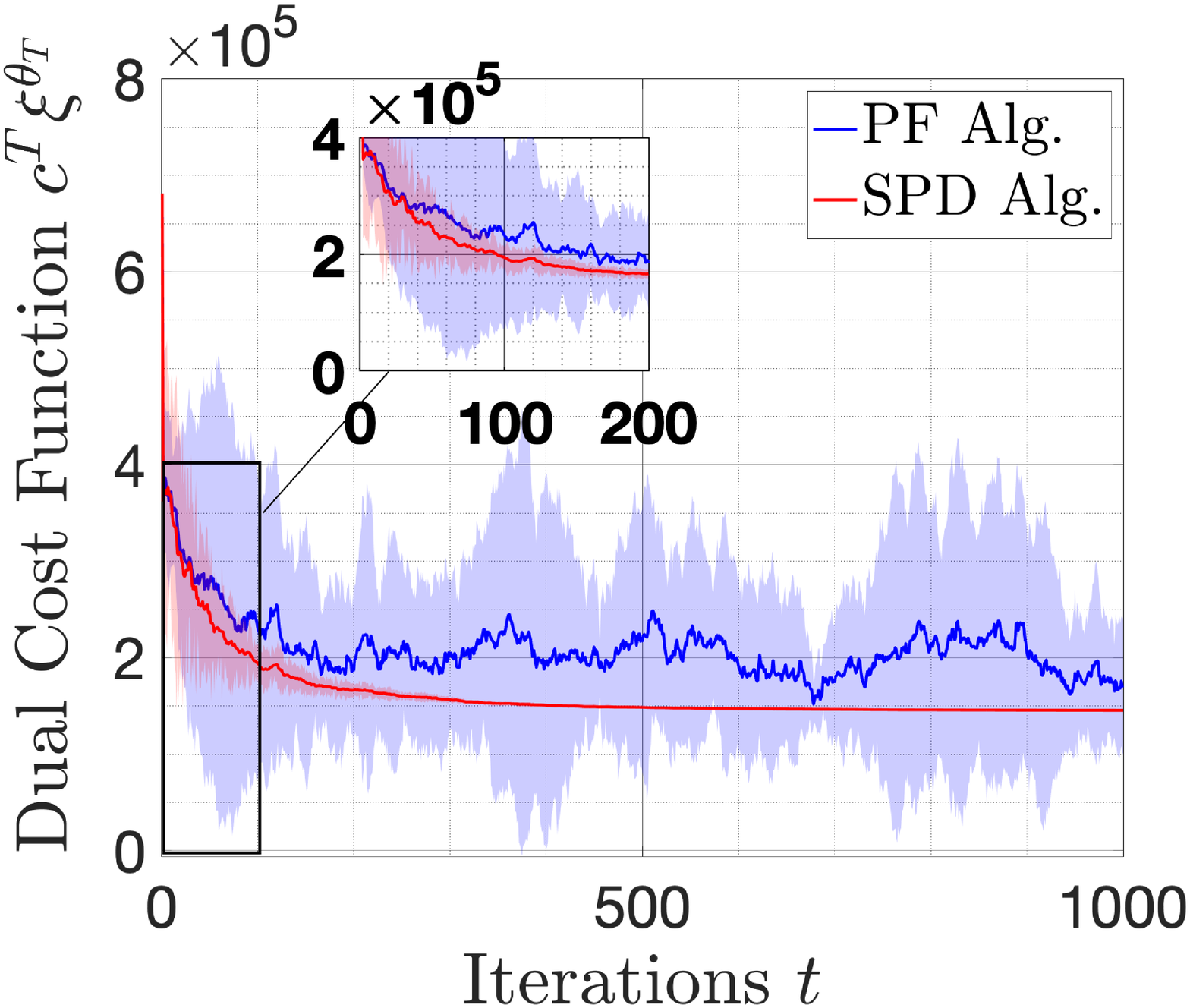}
		}
	\end{center}
\vspace{5mm}
	\caption{\footnotesize{Left: the mixing coefficients $\bm{\theta}_{t}=(\bm{\theta}_{t,1},\bm{\theta}_{t,2})$ during $t\in [T]$, $T=1000$ iterations using the stochastic primal-dual algorithm (Alg. \ref{Alg:1}) and the penalty function method of Banijamali \textit{et al.} \cite{banijamali2018optimizing}. Right: The dual cost function $\bm{c}^{T}\bm{\xi}^{\widehat{\bm{\theta}}_{T}}$. Each curve is obtained by averaging across 20 trials. Blue and red hulls around each curve show the 95\% confidence interval. 
			. } }
	\label{Fig:2}
\end{figure}

\subsection{Simulation Results}

We compare the performance of Algorithm \ref{Alg:1} with that of the penalty function method proposed by Banijamali \textit{et al.}. \cite{banijamali2018optimizing} to computing the mixture policies. Notice that each iteration of the penalty function method in \cite{banijamali2018optimizing} requires a projection onto the intersection of the norm ball and the hyperplane  $\bm{\Theta}_{R}\cap \bm{\Theta}_{2}$, which we implement using the $\tt{CVX}$ \cite{cvx} toolbox. In both algorithms, the same initial conditions and step-size $\eta_{t}=1/T=0.001$ are employed. Further, the size of the policy class in both algorithms is controlled by letting $R=100$, and using the $\ell_{2}$-norm. In addition, in the implementation of Algorithm \ref{Alg:1}, we employ the Bregman divergence specified in Eq. \eqref{Eq:Hellinger}, as well as the formula given for $z_{t}$ in Eq. \eqref{Eq:zt}. 

Figure \ref{Fig:2}-(a) depicts the mixing coefficients associated with the policies specified in Eq. \eqref{Eq:Policies_12}. In Figure \ref{Fig:2}-(b), we depict the cost $\bm{c}^{T}\bm{\xi}^{\widehat{\bm{\theta}}_{T}}$ of the dual ALP attained by Algorithm \ref{Alg:1} and the penalty function method. Clearly, our proposed method achieves a smaller cost compared to the penalty function method. The curves are obtained by averaging across 20 trials. The blue and red hulls around each curve are the 95\% confidence intervals. Since the confidence intervals associated with PD algorithm are small, we have magnified certain regions of each plot to better illustrate the red hulls. We observe that the primal-dual solutions have a smaller deviation across different trials compared to the penalty function method. Moreover, the computational complexity of each step of the penalty function in Banijamali \textit{et al.} \cite{banijamali2018optimizing} is significantly higher compared to Algorithm \ref{Alg:1} due to the projection step which requires a separate minimization problem to be solved in each round.

%

\section{Concluding Remarks}
\label{Section:Conclusion and Future Works}

We have studied the problem of learning control policies from a restricted mixture policy class, characterized by the convex hull of a set of known base policies. In particular, we considered a computation model where the state transition probability matrix is known, whereas the cost function is unknown. At each iteration, the agent takes an action at a particular state, and receives the associated cost from a sampling oracle. Subsequently, a new state is reached according to the transition probability matrix.

To compute the coefficients of the mixture policy of this model, we recast the problem as an approximate linear programming formulation of MDPs, where the feature matrix is characterized by the occupation measures of the base policies. Subsequently, we proposed a stochastic regularized primal-dual method to solve the corresponding ALP. The main contribution of the proposed algorithm in this paper is the simple projection step onto the hyper-plane of the feature vectors.

We analyzed the expanded efficiency of the stochastic primal-dual algorithm, and prove that it scales polynomially in the number of base policies. Furthermore, we demonstrated the performance of the proposed stochastic primal-dual algorithm to compute efficient mixture policies for queuing problems involving a single queue model as well as a four-dimensional queuing network. Compared to the penalty function method \cite{banijamali2018optimizing}, the proposed algorithm in this paper achieves better efficiencies.

\bibliographystyle{IEEEtran}
\bibliography{Masoud}
\appendix
	
\section{}	

\subsection{Proof of Proposition \ref{Prop:1}}
\label{Appendix:Proof_of_Proposition}

The proof of the identity \eqref{Eq:GLC2} is by direct computation:
\begin{align}
\nonumber
D_{\phi}(\bm{\theta};\bm{\theta}_{t})&+D_{\phi}(\bm{\theta}_{t};\widetilde{\bm{\theta}}_{t})=\phi(\bm{\theta})-\phi(\bm{\theta}_{t})-\langle \nabla\phi(\bm{\theta}_{t}), \bm{\theta}- \bm{\theta}_{t}\rangle\\ \nonumber
&+\phi(\bm{\theta}_{t})-\phi(\widetilde{\bm{\theta}}_{t})-\langle \nabla\phi(\widetilde{\bm{\theta}}_{t}), \bm{\theta}- \bm{\theta}_{t}\rangle\\ \nonumber
&=\phi(\bm{\theta})-\phi(\widetilde{\bm{\theta}}_{t})-\langle \nabla\phi(\widetilde{\bm{\theta}}_{t}),\bm{\theta}-\widetilde{\bm{\theta}}_{t} \rangle+\langle \nabla\phi(\widetilde{\bm{\theta}}_{t}),\bm{\theta}-\widetilde{\bm{\theta}}_{t} \rangle\\ \nonumber &
-\langle \nabla\phi(\bm{\theta}_{t}), \bm{\theta}- \bm{\theta}_{t}\rangle-\langle \nabla\phi(\widetilde{\bm{\theta}}_{t}), \bm{\theta}- \bm{\theta}_{t}\rangle\\ \nonumber
&=D_{\phi}(\bm{\theta};\widetilde{\bm{\theta}})+\langle \nabla\phi(\widetilde{\bm{\theta}}_{t}),\bm{\theta}-\bm{\theta}_{t} \rangle-\langle \nabla\phi(\bm{\theta}_{t}),\bm{\theta}-\bm{\theta}_{t} \rangle\\ \nonumber
&=D_{\phi}(\bm{\theta};\widetilde{\bm{\theta}}_{t})+\langle \nabla \phi(\bm{\theta}_{t})-\nabla \phi(\widetilde{\bm{\theta}}_{t}),\bm{\theta}-\bm{\theta}_{i}^{t} \rangle. \qedb
\end{align}

\subsection{Proof of Lemma \ref{Lemma:1}}
\label{Appendix:Proof_of_Lemma:1}
Let $\mathcal{F}_{t}$ denotes the $\sigma$-field containing all the randomness of the stochastic oracle up to time $t$. Thus, $\bm{\theta}_{t}$ is $\mathcal{F}_{t}$-measurable. Then,
\begin{align}
\expect[\langle \bm{e}_{t},\bm{\theta}_{t}-\bm{\theta} \rangle|\mathcal{F}_{t}]&=\langle \expect[\bm{e}_{t}|\mathcal{F}_{t}],\bm{\theta}_{t}-\bm{\theta}\rangle \label{Eq:zero_mean_Gaussian}
=0.
\end{align}
Moreover, the following upper bound holds due to the H\"{o}lder inequality,
\begin{align}
\nonumber
|\langle \bm{e}_{t},\bm{\theta}_{t}-\bm{\theta} \rangle| &\leq \|\bm{e}_{t}\|_{\ast}\cdot \|\bm{\theta}_{t}-\bm{\theta}\|\\
&\leq 2R\cdot \|\bm{e}_{t}\|_{\ast},
\end{align}
where the last inequality follows from the fact that $\bm{\theta}_{t},\bm{\theta}\in \bm{\Theta}_{R}^{d}$. In addition, we recall the definition of the gradient estimation error $\bm{e}_{t}$ from Eq. \eqref{Eq:Estimation_Error}
\small\begin{align}
\nonumber
\|\bm{e}_{t}\|_{\ast}&=\|\nabla_{\bm{\theta}}\widehat{\mathcal{L}}_{\bm{\Psi}}(\bm{\theta}_{t},\bm{\lambda}_{t})-\nabla_{\bm{\theta}}\mathcal{L}_{\bm{\Psi}}(\bm{\theta}_{t},\bm{\lambda}_{t})\|_{\ast}\\ \nonumber
&\leq \| \nabla_{\bm{\theta}}\widehat{\mathcal{L}}_{\bm{\Psi}}(\bm{\theta}_{t},\bm{\lambda}_{t})\|_{\ast}+\|\nabla_{\bm{\theta}}\mathcal{L}_{\bm{\Psi}}(\bm{\theta}_{t},\bm{\lambda}_{t}) \|_{\ast} \\ \nonumber
&= (|c_{a}(s_{t})|+|\lambda_{t,a_{t}}(s_{t})|) \left\|\dfrac{\bm{\Psi}_{a_{t}}(s_{t})}{{1\over d}\sum_{i=1}^{d}\psi_{a_{t}}^{i}(s_{t})}\right\|_{\ast}+\left\|\bm{c}^{T}\bm{\Psi}+\bm{\lambda}_{t}^{T}\bm{\Psi}\right\|_{\ast}\\ \label{Eq:Ce}
&\leq A(c_{\max}+G_{\beta})+(\sqrt{nm}c_{\max}+G_{\beta})\|\bm{\Psi}\|_{\ast,2}\df C_{e},
\end{align}\normalsize
where in the last step, we used the definition of the constant $A>0$ in Eq. \eqref{Constant:A}. 

The random variable $\langle \bm{e}_{t},\bm{\theta}_{t}-\bm{\theta} \rangle$ is (conditionally) zero-mean due to Eq. \eqref{Eq:zero_mean_Gaussian} and is bounded from above by $2RC_{e}$. Therefore, based on the characterization of the Orlicz norm in Definition \ref{Def:Orlicz}, we have
\begin{align}
\expect[\exp(| \langle\bm{e}_{t},\bm{\theta}_{t}-\bm{\theta}\rangle|^{2}/8R^{2}C_{e}^{2})]\leq \exp(1/2)\leq 2.
\end{align}
Hence, $\|\langle\bm{e}_{t},\bm{\theta}_{t}-\bm{\theta}\rangle \|_{\psi_{2}}\leq 8R^{2}C_{e}^{2}$. The following lemma state that the sum of sub-exponential random variables is also a sub-exponential random variable \textsc{\cite{wainwright2015basic}}:
\begin{lemma}\ \textsc{(Sum of Sub-Gaussian and Sub-Exponential r.v.'s)}
	\label{Lemma:Sum_Sub_Gaussian}	
	There exists a universal constant $C_{s}$, such that the sum of two dependent sub-Gaussian random variables with parameters $K_{1}$ and $K_{2}$ is $C_{s}(K_{1} +K_{2})$-sub-Gaussian, and the sum of two de- pendent sub-exponential random variables with parameters $K_{1}$ and $K_{2}$ is $C_{s}(K_{1}+K_{2})$-sub-exponential. $\hfill \square$
\end{lemma}	

The proof of lemma \ref{Lemma:Sum_Sub_Gaussian} follows directly from the equivalent form of definition of sub-Gaussian and sub- exponential random variables using Orlicz norms.

We now use the following martingale-based bound for variables with conditionally sub-Gaussian
behavior which is of the Azuma-H\"{o}effding type (cf. \cite{azuma1967weighted,hoeffding1963probability}):
\begin{lemma}
	\textsc{(Azuma-H\"{o}effding Inequality)}
	Let $X_{t}$ be a martingale difference sequence adapted to the filtration $\mathscr{F}_{t}$ and assume each $X_{t}$ is conditionally sub-Gaussian with the Orlicz norm of $\|X_{t}\|_{\psi_{2}}\leq \sigma_{t}$, meaning that 
	\begin{align}
	\expect\left[\exp(|X_{t}|^{2}/\sigma_{t}^{2})|\mathcal{F}_{t}\right]\leq 2. 
	\end{align}
	Then, for all $\varepsilon>0$,
	\begin{align}
	\label{Eq:Azuma_Hoeffding}
	\prob\left[\left|\sum_{t=0}^{T-1}X_{t}\geq \varepsilon\right|\right]\leq 2\exp\left(-\dfrac{c\cdot\varepsilon^{2}}{\sum_{t=0}^{T-1}\sigma_{t}} \right).\\
	\qedw
	\end{align}
\end{lemma}
Applying the Azuma-H\"{o}effding inequality \eqref{Eq:Azuma_Hoeffding} to the sum $\sum_{t=0}^{T-1}\eta_{t}\langle \bm{e}_{t},\bm{\theta}_{t}-\bm{\theta}\rangle$ yields the  concentration bound in Eq. \eqref{Eq:CON_11}. 

To prove the concentration bound in Eq. , we first state the following lemma:
\begin{lemma}\ \textsc{(Squared Norm of Sub-Gaussian Random Variable)}
	\label{Lemma:3}
	Assume that $\bm{X}\in \real^{d}$ is zero mean, and is a sub-Gaussian random variable with the Orlicz norm $\|\bm{X}\|_{\psi_{2}}\leq K$. Then, the random variable $Z=\|\bm{X}\|_{2}^{2}$ is sub-exponential with the variable $\|Z\|_{\psi_{1}}\leq (\sqrt{\log(2)}/2) K^{2}$. 
\end{lemma}
The proof of Lemma \ref{Lemma:3} is defered to Appendix \ref{Appendix:Proof_Lemma:3}. 

Now, we recall from Eq. \eqref{Eq:Estimation_Error} that $\bm{e}_{t}$ is (conditionally) zero mean, \textit{i.e.}, $\expect[\bm{e}_{t}|\mathcal{F}_{t}]=0$. Moreover, as shown in Appendix \ref{Appendix:Proof_of_Lemma:1}, the estimation error has a bounded norm $\|\bm{e}_{t}\|_{\ast}\leq  C_{e}$.  Therefore, for any vector on the sphere $\bm{v}\in \mathrm{S}^{n-1}$, the following inequalities hold
\begin{align}
\nonumber
\exp[|\langle \bm{e}_{t},\bm{v}\rangle|^{2}/K^{2}]&\stackrel{\rm{(a)}}{\leq} \exp[\|\bm{e}_{t}\|^{2}_{\ast}\cdot \|\bm{v}\|^{2}/K^{2}]\\ \label{Eq:We_see}
&\stackrel{\rm{(b)}}{\leq} \exp[C^{2}(n)C_{e}^{2}\cdot \|\bm{v}\|^{2}_{2}/K^{2}].
\end{align}
where $\rm{(a)}$ follows form the Cauchy-Schwarz inequality, and in $\rm{(b)}$, $C(n)$ is a constant such that $\|\bm{v}\|\leq C_{n}\|\bm{v}\|_{2}$.  From Eq. \eqref{Eq:We_see}, we have that if $K=\sqrt{2}C(n)C_{e}$, then $\exp[|\langle \bm{e}_{t},\bm{v}\rangle|^{2}/K^{2}]\leq \exp(1/2)\leq 2$ for all $\bm{v}\in \mathrm{S}^{n-1}$.  Thus, the vector of the estimation error $\bm{e}_{t}$ is sub-Gaussian with the Orlicz norm of 
\begin{align}
\|\bm{e}_{t}\|_{\psi_{2}}\leq \sqrt{2}C(n)C_{e}. 
\end{align}
Leveraging Lemma \ref{Lemma:3}, it follows that $\|\bm{e}_{t}\|_{\ast}^{2}$ is a sub-exponential random variable with the Orlicz norm of $\| \|\bm{e}_{t}\|_{\ast}^{2} \|_{\psi_{1}}\leq \sqrt{{\log(2)\over 2}}C^{2}(n)C_{e}^{2}\leq C^{2}(n)C_{e}^{2}$. 

\begin{lemma}\ \textsc{(Bernstein inequality for sub-exponential random variables \cite{wainwright2015basic})}
	\label{Lemma:Bernstein inequality}
	Let $X_{0}, \cdots,X_{T-1}$ be independent sub-exponential random variables with $\|X_{t}\|_{\psi_{1}}\leq b_{t}$, and define $S_{T} = \sum_{i=0}^{T-1}X_{i}-\expect[X_{i}]$. Then there exists a universal constant $c>0$ such that, for all $t>0$,
	\begin{align}
	\label{Eq:Bernstein_Inequality}
	\prob(S_{T}\geq \delta)\leq \exp\left(-c\cdot \min\left\{{\delta^{2}\over \|\bm{b}\|_{2}^{2}},{\delta\over \|\bm{b}\|_{\infty}}\right\}\right),
	\end{align}
	where $\bm{b}=(b_{0},\cdots,b_{T-1})$.
	
	\hfill $\square$
\end{lemma}

We now apply the Bernstein concentration bound in Eq. \eqref{Eq:Bernstein_Inequality} of Lemma \eqref{Lemma:Bernstein inequality} to the sum of variances of the estimation error vector and invoke:
\small{\begin{align*}
&\prob\Bigg[\sum_{t=0}^{T-1}\eta_{t}^{2}(\|\bm{e}_{t}\|_{2}^{2}-\expect[\|\bm{e}_{t}\|_{2}^{2}|\mathcal{F}_{t}])\geq \delta \Bigg]\leq \exp\left(-c\cdot \min\left\{ \dfrac{\delta^{2}}{C_{e}^{2}C^{2}(n)\sum_{t=0}^{T-1}\eta_{t}^{4}},\dfrac{\delta}{C_{e}C(n)\sum_{t=0}^{T-1}\eta_{t}^{2}} \right\} \right).
\end{align*}}\normalsize

\subsection{Proof of Lemma \ref{Lemma:3}}
\label{Appendix:Proof_Lemma:3}

We can assume $\tau^{2}=1$ by scale invariance. Let $\bm{W}\sim \mathsf{N}(0,\bm{I}_{d\times d})$ be independent of $\bm{X}$. It is known that $\expect[e^{\lambda Z^{2}}]=\dfrac{1}{\sqrt{1-2\lambda}}, Z\sim \mathsf{N}(0,1)$, and $\lambda<1/2$. So, we have that
\begin{align*}
\expect[\lambda\|\bm{W}\|_{2}^{2}]=\prod_{i=1}^{d}\expect[\lambda W_{i}^{2}]=\dfrac{1}{(1-2\lambda)^{d/2}},
\end{align*}
for any $\lambda<1/2$. Now, for any $\lambda>0$, we evaluate the quantity $\expect[\sqrt{2\lambda}\langle \bm{X},\bm{W} \rangle]$ in two ways by conditioning on $\bm{X}$ and then $\bm{W}$. We have
\begin{align}
\nonumber 
\expect[\exp(\sqrt{2\lambda}\langle \bm{X},\bm{W} \rangle)]&=\expect\left[\expect\left[\exp(\sqrt{2\lambda}\langle \bm{X},\bm{W} \rangle)\right]|\bm{X}\right]\\ \nonumber
&= \expect\left[\exp(\|\sqrt{2\lambda}\bm{X}\|_{2}^{2}/2)\right]\\ \label{Eq:COMBI_1}
&=\expect[e^{\lambda\|\bm{X}\|_{2}^{2}}].
\end{align}
On the other hand, we have
\begin{align}
\nonumber
\expect[\exp(\sqrt{2\lambda}\langle \bm{X},\bm{W} \rangle)]&=\expect\left[\expect\left[\exp(\sqrt{2\lambda}\langle \bm{X},\bm{W} \rangle)\right]\Bigg|\bm{W}\right]\\ \nonumber
&\leq \expect\left[e^{\|\sqrt{2\lambda}\bm{X}\|_{2}/2}\right]\\  \nonumber
&=\expect\left[e^{\lambda\|\bm{W}\|_{2}^{2}}\right]\\  \label{Eq:COMBI_2}
&=\dfrac{1}{(1-2\lambda)^{d/2}}.
\end{align}
Now, by Jensen's inequality $\expect[e^{\lambda \|\bm{X}\|_{2}^{2}}]\geq e^{\lambda\expect[\|\bm{X}\|_{2}^{2}]}$. Further, putting together Eqs. \eqref{Eq:COMBI_1} and \eqref{Eq:COMBI_2}, yields
\begin{align*}
\expect\left[e^{\lambda\big( \big|\|\bm{X}\|_{2}^{2}-\expect[\|\bm{X}\|_{2}^{2}]\big|\big)}\right]=\expect\left[e^{\lambda (\|\bm{X}\|_{2}^{2}-\expect[\|\bm{X}\|_{2}^{2}])}\right]&\leq \expect\left[e^{\lambda\|\bm{X}\|_{2}^{2}}\right]\\
&\leq \dfrac{1}{(1-2\lambda)^{d/2}}\\
&\leq \exp(4\lambda^{2}).
\end{align*}  
Letting $\lambda=\sqrt{\log(2)}/2$ then yields
\begin{align}
\expect\left[e^{\sqrt{\log(2)}(\|\bm{X}\|_{2}^{2}-\expect[\|\bm{X}\|_{2}^{2}])/2}\right]\leq \exp(\log(2))= 2.
\end{align}
Therefore, $\|\|\bm{X}\|_{2}^{2}\|_{\psi_{1}}\leq \sqrt{\log(2)}/2$ by the definition of the Orlicz norm in Definition \ref{Def:Orlicz}.

\bibliographystyle{plain}
\bibliography{Masoud}

%


%
%
%

\end{document}